\documentclass{article}

    \PassOptionsToPackage{numbers, compress}{natbib}
     \usepackage[final]{neurips_2020}

\usepackage{natbib}
\pdfoutput=1

\usepackage[hidelinks]{tre_packages}

\graphicspath{{./figs/}}

\title{Telescoping Density-Ratio Estimation}

\author{%
  Benjamin Rhodes \\
  School of Informatics\\
  University of Edinburgh\\
  \texttt{ben.rhodes@ed.ac.uk} \\
  \And
    Kai Xu \\
  School of Informatics \\
  University of Edinburgh \\
  \texttt{kai.xu@ed.ac.uk} \\
  \And
  Michael U. Gutmann \\
  School of Informatics \\
  University of Edinburgh \\
  \texttt{michael.gutmann@ed.ac.uk} \\
}

\begin{document}

\maketitle

\begin{abstract}
    Density-ratio estimation via classification is a cornerstone of unsupervised learning. It has provided the foundation for state-of-the-art methods in representation learning and generative modelling, with the number of use-cases continuing to proliferate. However, it suffers from a critical limitation: it fails to accurately estimate ratios $p/q$ for which the two densities differ significantly. Empirically, we find this occurs whenever the KL divergence between $p$ and $q$ exceeds tens of nats. To resolve this limitation, we introduce a new framework, telescoping density-ratio estimation (TRE), that enables the estimation of ratios between highly dissimilar densities in high-dimensional spaces. Our experiments demonstrate that TRE can yield substantial improvements over existing single-ratio methods for mutual information estimation, representation learning and energy-based modelling.
\end{abstract}

\section{Introduction}

Unsupervised learning via density-ratio estimation is a powerful paradigm in machine learning \cite{sugiyama2012density} that continues to be a source of major progress in the field. It consists of estimating the ratio $p/q$ from their samples without separately estimating the numerator and denominator. A common way to achieve this is to train a neural network classifier to distinguish between the two sets of samples, since for many loss functions the ratio $p/q$ can be extracted from the optimal classifier \cite{sugiyama2012density, gutmann2012bregman, menon2016linking}. This discriminative approach has been leveraged in diverse areas such as covariate shift adaptation \cite{sugiyama2008direct, tsuboi2009direct}, energy-based modelling \cite{Gutmann2012a, ceylan2018conditional, rhodes2019variational, tu2007learning, lazarow2017introspective, grover2018boosted}, generative adversarial networks \cite{goodfellow2014generative, nowozin2016f, mohamed2016learning}, bias correction for generative models \cite{grover2019bias, grover2019fair}, likelihood-free inference \cite{pham2014note, thomas2016likelihood, dinev2018dynamic, durkan2020contrastive}, mutual-information estimation \cite{belghazi2018mutual}, representation learning \cite{hyvarinen2016unsupervised, hyvarinen2019nonlinear,  oord2018representation, henaff2019data, hjelm2018learning}, Bayesian experimental design \cite{kleinegesse2019efficient, kleinegesse2020bayesian} and off-policy reward estimation in reinforcement learning \cite{liu2018breaking}. Across this diverse set of applications, density-ratio based methods have consistently yielded state-of-the-art results.

\begin{figure}[t]
\centering
\begin{subfigure}{1.0\textwidth}
  \centering
%   \hspace{3mm}
%   {\Large \textbf{$\bf{\frac{p}{q}}$}}
%   \vspace*{0.4cm}
  \includegraphics[width=.99\linewidth]{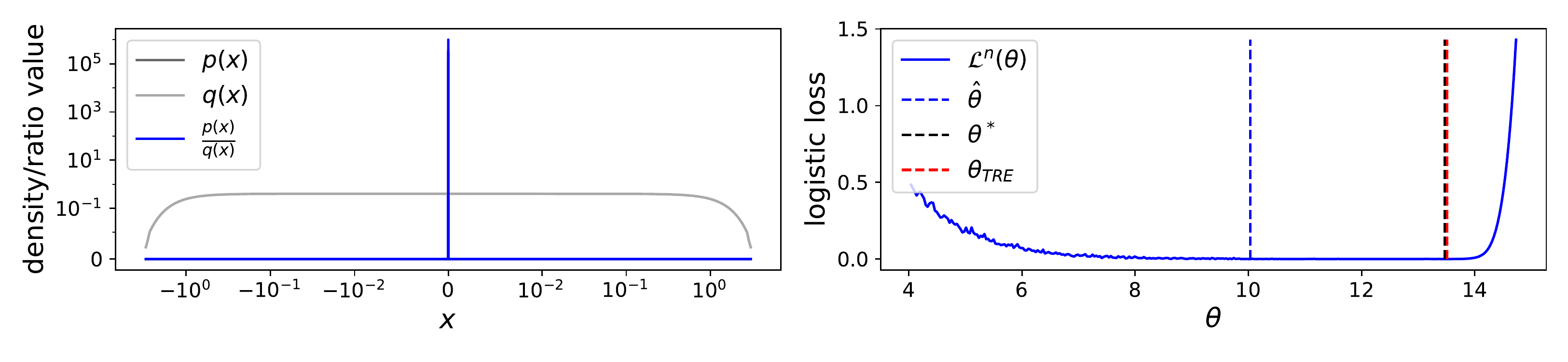}
  \vspace*{-0.3cm} % xlabel not visible for values < -0.3
  \caption{Density-ratio estimation between an extremely peaked Gaussian $p$ ($\sigma=10^{-6}$) and a broad Gaussian $q$ ($\sigma=1$) using a single-parameter quadratic classifier (as detailed in section \ref{sec: 1d peaked gauss experiment}). \textbf{Left}: A log-log scale plot of the densities and their ratio. Note that p(x) is not visible, since the ratio overlaps it. \textbf{Right}: the solid blue line is the finite-sample logistic loss (Eq.\ \ref{eq: finite logistic loss}) for 10,000 samples. Despite the large sample size, the minimiser (dotted blue line) is far from optimal (dotted black line). The dotted red line is the newly introduced TRE solution, which almost perfectly overlaps with the dotted black line.}
  \vspace*{0.2cm}
  \label{fig:1d single ratio subfig}
\end{subfigure}
\begin{subfigure}{1.0\textwidth}
  \centering
    \hspace{-0.4cm}
        \begin{tabular}{c c c c}
        \hspace{-12mm}
        \Large $\boldsymbol{\frac{p}{q}}$
        \hspace{4mm} \Large $\boldsymbol{=}$
        \hspace{3mm} \Large $\boldsymbol{\frac{p}{p_1}}$ &
        \hspace{-20mm} \Large $\boldsymbol{\times}$
        \hspace{12mm} \Large $\boldsymbol{\frac{p_1}{p_2}}$ &
        \hspace{-20mm} \Large $\boldsymbol{\times}$
        \hspace{12mm} \Large $\boldsymbol{\frac{p_2}{p_3}}$ &
        \hspace{-20mm} \Large $\boldsymbol{\times}$
        \hspace{12mm} \Large $\boldsymbol{\frac{p_3}{q}}$ \\
        \includegraphics[width=.24\linewidth]{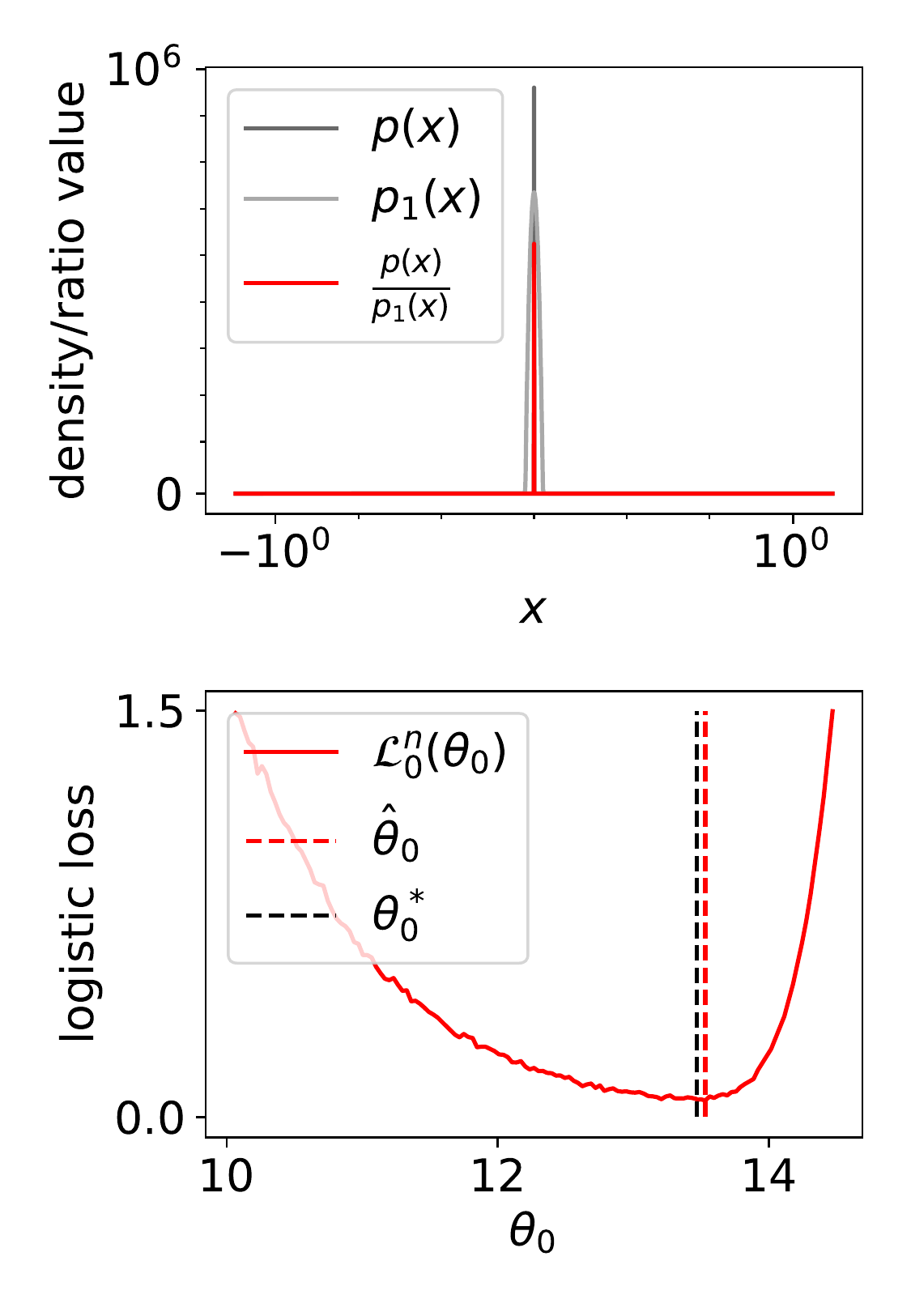} &
        \hspace{-0.5cm} \includegraphics[width=.24\linewidth]{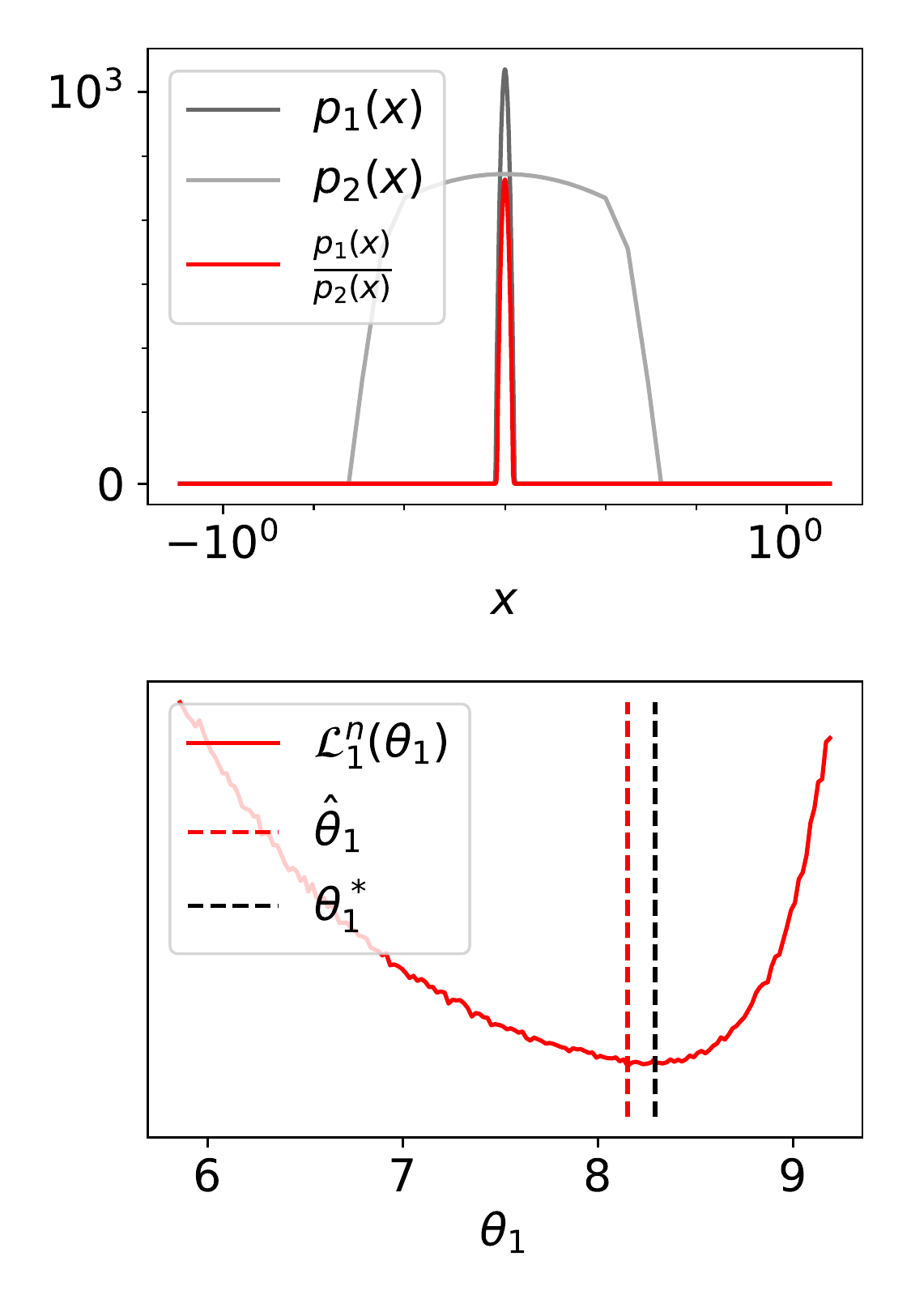} &
        \hspace{-0.5cm}
        \includegraphics[width=.24\linewidth]{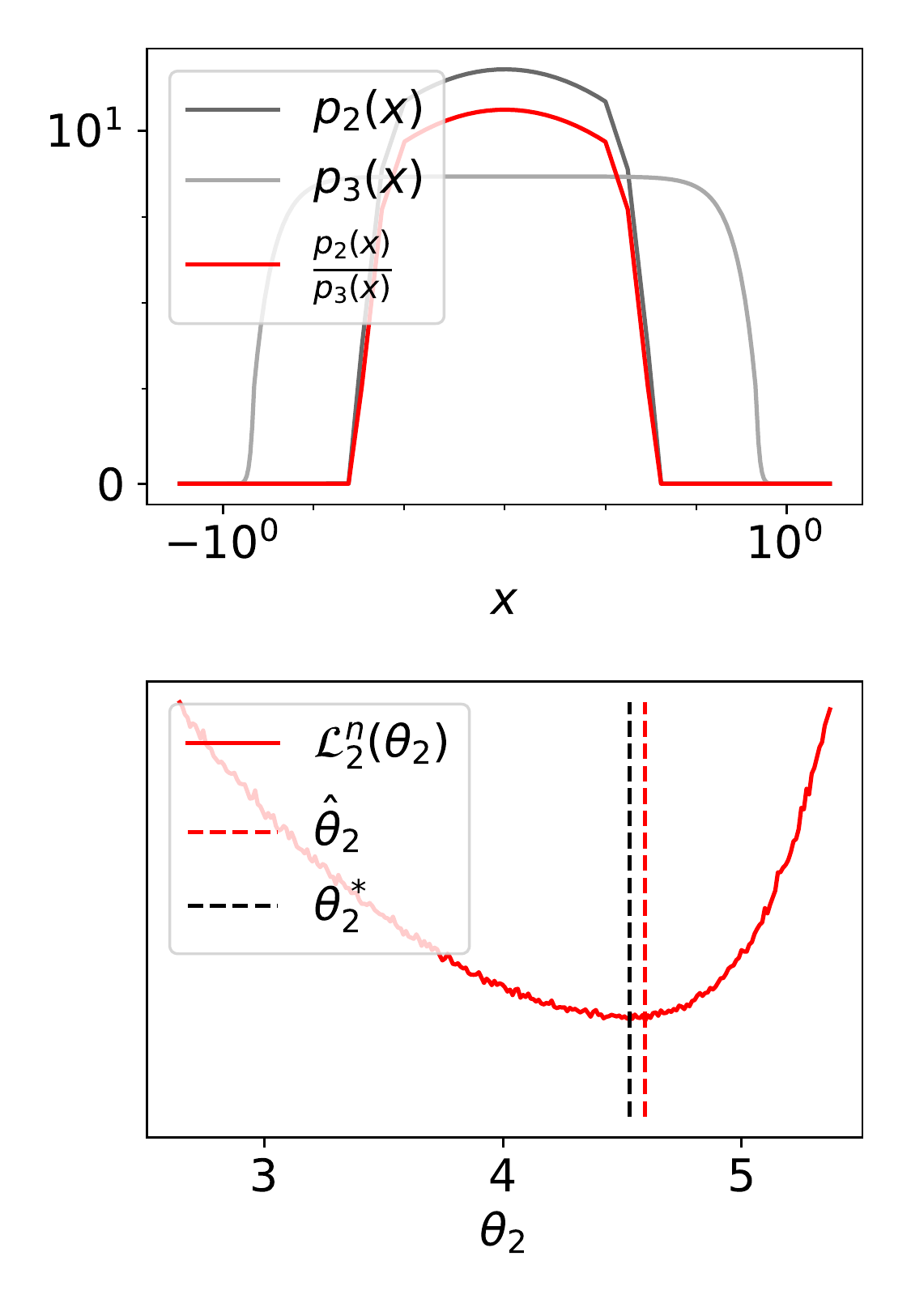} &
        \hspace{-0.5cm}
        \includegraphics[width=.24\linewidth]{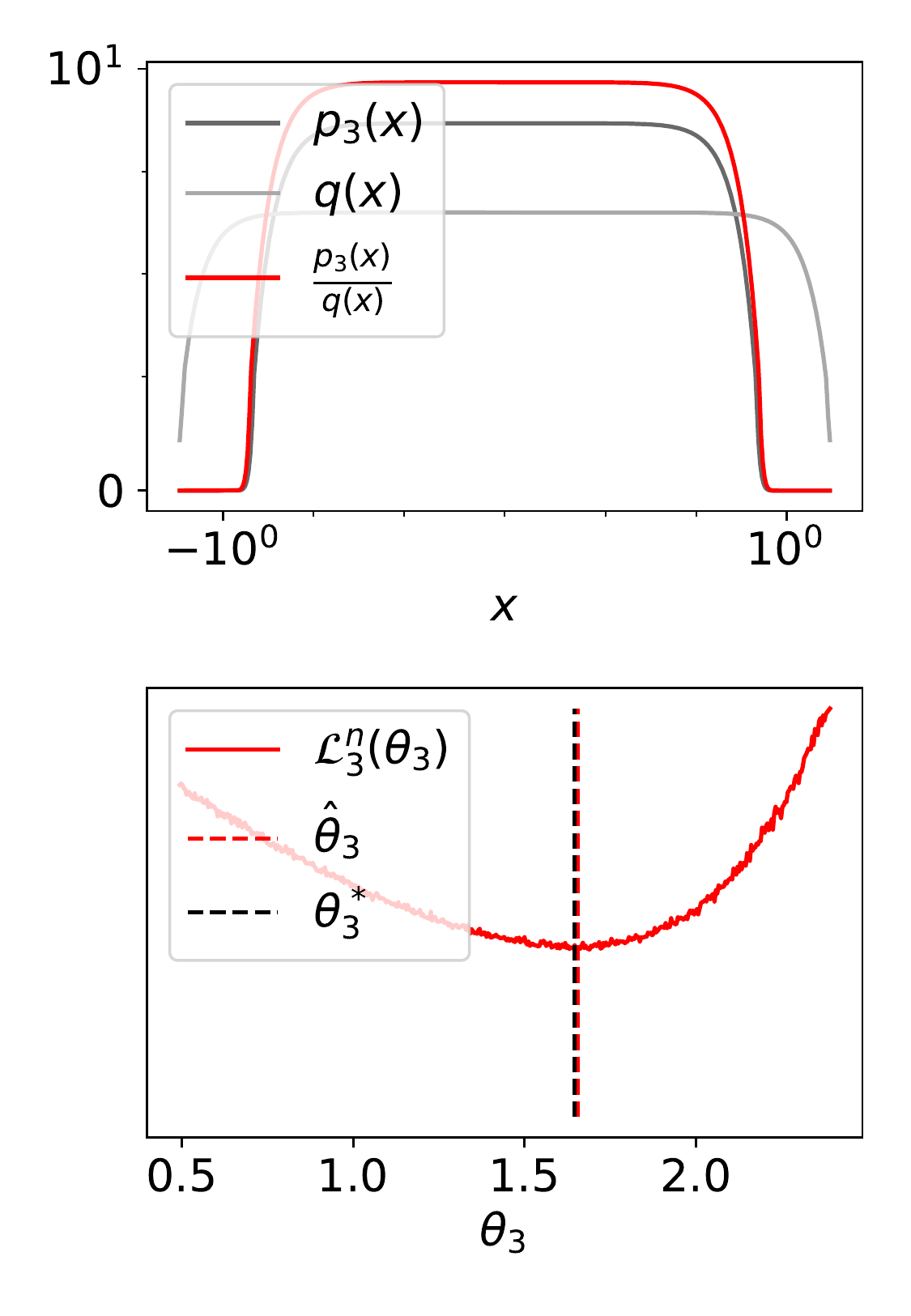}
    \end{tabular}
\vspace*{-0.3cm}
\caption{Telescoping density-ratio estimation applied to the problem in (a), using the same 10,000 samples from $p$ and $q$. \textbf{Top row}: a collection of ratios, where $p_1$, $p_2$ and $p_3$ are constructed by deterministically interpolating between samples from $p$ and $q$. \textbf{Bottom row}: the logistic loss function for each ratio estimation problem. Observe that the finite-sample minimisers of each objective (red dotted lines) are either close to or exactly overlapping their optima (black dotted lines). After estimating each ratio, we then combine them by taking their product.}
\label{fig:1d tre subfig}
\end{subfigure}
\caption{Illustration of standard density-ratio estimation vs.\ telescoping density-ratio estimation.}
\label{fig:1d single ratio vs tre}
\end{figure}

% Many existing loss functions for discriminative density-ratio estimation have a severe limitation however.
Despite the successes of discriminative density-ratio estimation, many existing loss functions share a severe limitation. Whenever the `gap' between $p$ and $q$ is large, the classifier can obtain almost perfect accuracy with a relatively poor estimate of the density ratio. We refer to this failure mode as the \emph{density-chasm problem}---see Figure \ref{fig:1d single ratio subfig} for an illustration. We observe empirically that the density-chasm problem manifests whenever the KL-divergence $\infdiv{KL}{p}{q}$ exceeds $\sim 20$ nats\footnote{`nat' being a unit of information measured using the natural logarithm (base $e$)}. This observation accords with recent findings in the mutual information literature regarding the limitations of density-ratio based estimators of the KL \cite{mcallester2018formal, poole2019variational, song2019understanding}. In high dimensions, it can easily occur that two densities $p$ and $q$ will have a KL-divergence measuring in the hundreds of nats, and so the ratio may be virtually intractable to estimate with existing techniques.
 
In this paper, we propose a new framework for estimating density-ratios that can overcome the density-chasm problem. Our solution uses a `divide-and-conquer' strategy composed of two steps. The first step is to gradually transport samples from $p$ to samples from $q$, creating a chain of intermediate datasets. We then estimate the density-ratio between consecutive datasets along this chain, as illustrated in the top row of Figure \ref{fig:1d tre subfig}. Unlike the original ratio $p/q$, these `chained ratios' can be accurately estimated via classification (see bottom row). Finally, we combine the chained ratios via a telescoping product to obtain an estimate of the original density-ratio $p/q$. Thus, we refer to the method as Telescoping density-Ratio Estimation (TRE).

We empirically demonstrate that TRE can accurately estimate density-ratios using deep neural networks on high-dimensional problems, significantly outperforming existing single-ratio methods. We show this for two important applications: representation learning via mutual information (MI) estimation and the learning of energy-based models (EBMs).

In the context of mutual information estimation, we show that TRE can accurately estimate large MI values of 30+ nats, which is recognised to be an outstanding problem in the literature \cite{poole2019variational}. However, obtaining accurate MI estimates is often not our \emph{sole} objective; we also care about learning representations from e.g.\ audio or image data that are useful for downstream tasks such as classification or clustering. To this end, our experimental results for representation learning confirm that TRE offers substantial gains over a range of existing single-ratio baselines.

In the context of energy-based modelling, we show that TRE can be viewed as an extension of noise-contrastive estimation \cite{Gutmann2012a} that more efficiently scales to high-dimensional data. Whilst energy-based modelling has been a topic of interest in the machine learning community for some time \cite{ruslan2009deep}, there has been a recent surge of interest, with a wave of new methods for learning deep EBMs in high dimensions \cite{du2019implicit, dai2019exponential, song2019generative, li2019annealed, grathwohl2020cutting, yu2020training}. These methods have shown promising results for image and 3D shape synthesis \cite{xie2018learning}, hybrid modelling \cite{grathwohl2019your}, and modelling of exchangeable data \cite{yang2020energy}.

However, many of these methods result in expensive/challenging optimisation problems, since they rely on approximate Markov chain Monte Carlo (MCMC) sampling during learning \cite{du2019implicit, grathwohl2019your, yu2020training}, or on adversarial optimisation \cite{dai2019exponential, grathwohl2020cutting, yu2020training}. In contrast, TRE requires no MCMC during learning and uses a well-defined, non-adversarial, objective function. Moreover, as we show in our mutual information experiments, TRE is applicable to discrete data, whereas all other recent EBM methods only work for continuous random variables. Applicability to discrete data makes TRE especially promising for domains such as natural language processing, where noise-contrastive estimation has been widely used \cite{mikolov2013distributed, kong2019mutual, bakhtin2020energy}.

\section{Discriminative ratio estimation and the density-chasm problem}
\label{sec: dre}
Suppose $p$ and $q$ are two densities for which we have samples, and that $q(\x) > 0$ whenever $p(\x) > 0$. We can estimate the density-ratio $r(\x) = p(\x)/q(\x)$ by training a classifier to distinguish samples from $p$ and $q$ \cite{hastie2009elements, sugiyama2012density, Gutmann2012a}. There are many choices for the loss function of the classifier \cite{sugiyama2012density, Pihlaja2010, gutmann2012bregman, menon2016linking, poole2019variational}, but in this paper we concentrate on the widely used logistic loss
\begin{align}
 \L(\thetab) = - \E_{\x_1 \sim p} \log \left( \frac{r(\x_1; \thetab)}{1 + r(\x_1; \thetab)}  \right) -
 \E_{\x_2 \sim q} \log \left( \frac{1}{1 + r(\x_2; \thetab)}  \right),
 \label{eq: logistic loss}
\end{align}
where $r(\x; \thetab)$ is a non-negative ratio estimating model. To enforce non-negativity, $r$ is typically expressed as the exponential of an unconstrained function such as a neural network. For a correctly specified model, the minimiser of this loss, $\thetab^*$, satisfies $r(\x; \thetab^*) = p(\x)/q(\x)$, without needing any normalisation constraints \cite{Gutmann2012a}. Other classification losses do not always have this self-normalising property, but only yield an estimate proportional to the true ratio---see e.g.\ \cite{poole2019variational}.

% \subsection{The density-chasm problem}
\subsection*{The density-chasm problem}
%\textbf{The density-chasm problem}

\begin{wrapfigure}{r}{0.3\linewidth}
\vspace{-1cm}
\centering
\includegraphics[width=0.99\linewidth]{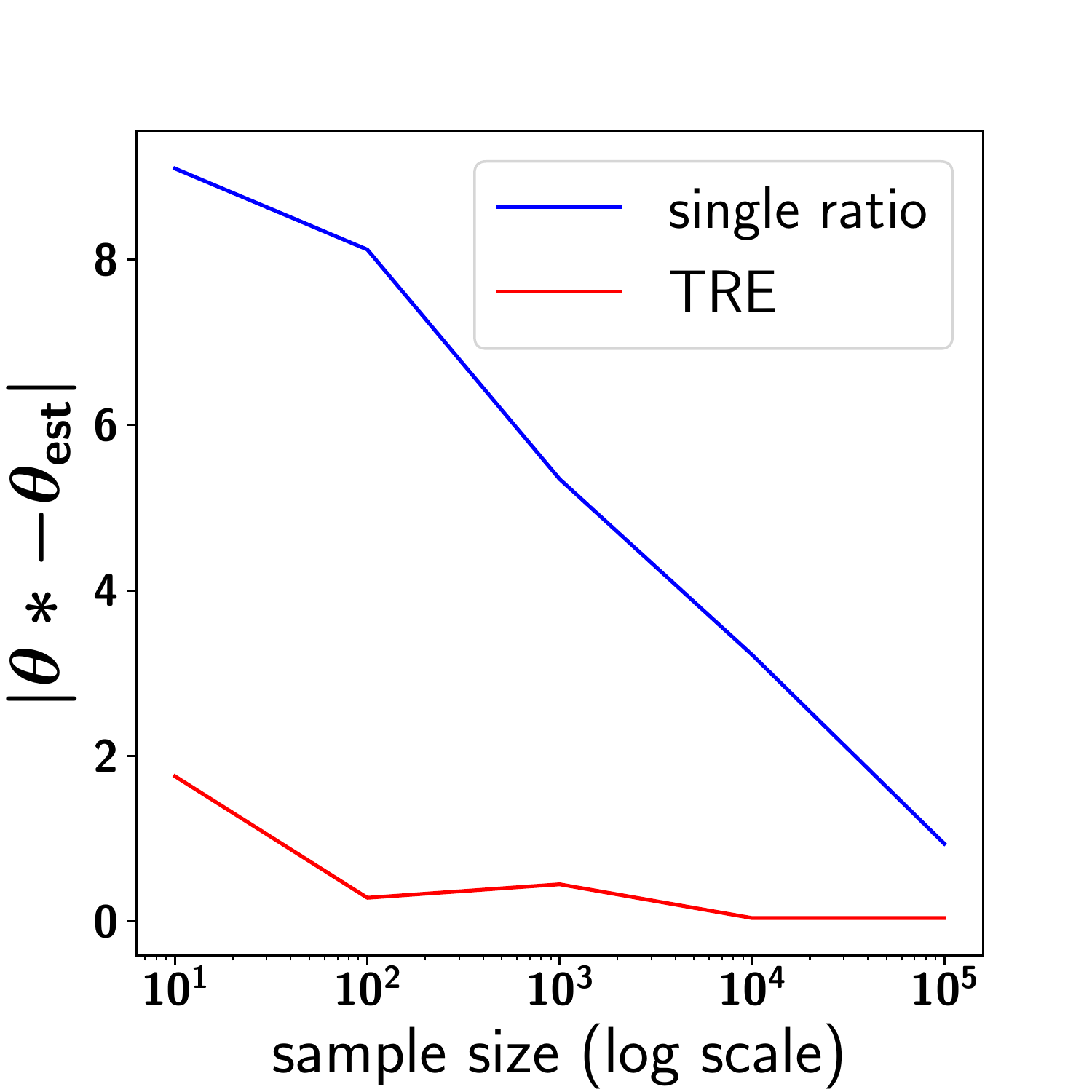} 
\caption{Sample efficiency curves for the experiment in Figure \ref{fig:1d single ratio vs tre}. Single ratio estimation can be extremely sample-inefficient.}
\label{fig:sample efficiency curves}
\vspace{-0.2cm}
\end{wrapfigure}
We experimentally find that density-ratio estimation via classification typically works well when $p$ and $q$ are `close' e.g.\ the KL divergence between them is less than $\sim 20$ nats. However, for sufficiently large gaps, which we refer to as \emph{density-chasms}, the ratio estimator is often severely inaccurate. This raises the obvious question: what is the cause of such inaccuracy?

There are many possible sources of error: the use of misspecified models, imperfect optimisation algorithms, and inaccuracy stemming from Monte Carlo approximations of the expectations in \eqref{eq: logistic loss}. We argue that this mundane final point---Monte Carlo error due to finite sample size---is actually sufficient for inducing the density-chasm problem. Figure \ref{fig:1d single ratio subfig} depicts a toy problem for which the model is well-specified, and because it is 1-dimensional (w.r.t.\ $\theta$), optimisation is straightforward using grid-search. And yet, if we use a sample size of $n=10,000$ and minimise the finite-sample loss
% \vspace{-5mm}
\begin{align}
 \L^n(\theta) &= \sum_{i=1}^{n} - \log \left( \frac{r(x^i_1; \theta)}{1 + r(x^i_1; \theta)}  \right) - \log \left( \frac{1}{1 + r(x^i_2; \theta)}  \right), & x^i_1 \sim p, \ x^i_2 \sim q,
 \label{eq: finite logistic loss}
\end{align}
we obtain an estimate $\hat{\theta}$ that is far from the asymptotic minimiser $\theta^* = \argminempty\L(\theta)$. Repeating this same experiment for different sample sizes, we can empirically measure the method's sample efficiency, which is plotted as the blue curve in Figure \ref{fig:sample efficiency curves}. For the regime plotted, we see that an exponential increase in sample size only yields a linear decrease in estimation error. This empirical result is concordant with theoretical findings that density-ratio based lower bounds on KL divergences are only tight for sample sizes exponential in the the number of nats \cite{mcallester2018formal}.

Whilst we focus on the logistic loss, we believe the density chasm problem is a broader phenomenon. As shown in the appendix, the issues identified in Figure 1 and the sample inefficiency seen in Figure 2 also occur for other commonly used discriminative loss functions.

Thus, when faced with the density-chasm problem, simply increasing the sample size is a highly inefficient solution and not always possible in practice. This begs the question: is there a more intelligent way of using a fixed set of samples from $p$ and $q$ to estimate the ratio?

\section{Telescoping density-ratio estimation}
\label{sec: tre}
We introduce a new framework for estimating density-ratios $p/q$ that can overcome the density-chasm problem in a \emph{sample-efficient} manner. Intuitively, the density-chasm problem arises whenever classifying between $p$ and $q$ is `too easy'. This suggests that it may be fruitful to decompose the task into a collection of harder sub-tasks.

For convenience, we make the notational switch $p \equiv p_0, \ \ q \equiv p_m$ (which we will keep going forward), and expand the ratio via a telescoping product
\begin{align}
     \frac{p_0(\x)}{p_m(\x)} = 
     \frac{p_0(\x)}{p_1(\x)} \frac{p_1(\x)}{p_2(\x)}  \ldots \frac{p_{m-2}(\x)}{p_{m-1}(\x)} \frac{p_{m-1}(\x)}{p_m(\x)},
\end{align}
where, ideally, each $p_k$ is chosen such that a classifier cannot easily distinguish it from its two neighbouring densities. Instead of attempting to build one large `bridge' (i.e.\ density-ratio) across the density-chasm, we propose to build many small bridges between intermediate `waymark' distributions. The two key components of the method are therefore:
\begin{enumerate}
\item \textbf{Waymark creation.} We require a method for \emph{gradually} transporting samples $\{\x^1_0, \ldots , \x^n_0\}$ from $p_0$ to samples $\{ \x^1_m, \ldots , \x^n_m \}$ from $p_m$. At each step in the transportation, we obtain a new dataset $\{ \x^1_k, \ldots , \x^n_k \}$ where $k \in \{0, \ldots m\}$. Each intermediate dataset can be thought of as samples from an implicit distribution $p_k$, which we refer to as a \emph{waymark} distribution.

\item \textbf{Bridge-building:} A method for learning a set of parametrised density-ratios between consecutive pairs of waymarks $r_k(\x ; \thetab_k) \approx p_k(\x)/p_{k+1}(\x)$ for $k = 0, \ldots, m-1$, where each bridge $r_k$ is a non-negative function. We refer to these ratio estimating models as \emph{bridges}. Note that the parameters of the bridges, $ \{\thetab_k\}_{k=0}^{m-1}$, can be totally independent or they can be partially shared.
\end{enumerate}
An estimate of the original ratio is then given by the product of the bridges
\begin{align}
    r(\x; \thetab) &= 
    \prod_{k=0}^{m-1} r_k(\x ; \thetab_k) 
    \approx \prod_{k=0}^{m-1}  \frac{p_k(\x)}{p_{k+1}(\x)}
    =  \frac{p_0(\x)}{p_m(\x)} \label{eq:telescope product}, 
\end{align}
where $\thetab$ is the concatenation of all $\thetab_k$ vectors. Because of the telescoping product in \eqref{eq:telescope product}, we refer to the method as Telescoping density-Ratio Estimation (TRE).

% TRE has conceptual ties with a range of methods in optimisation, statistical physics and machine learning that leverage sequences of intermediate distributions, typically between a complex density $p$ and a simple tractable density $q$. Of particular note are the methods of Simulated Annealing \cite{kirkpatrick1983sim}, Bridge Sampling \& Path Sampling \cite{gelman1998path} and Annealed Importance Sampling (AIS) \cite{neal2001annealed}. The connection to AIS is especially strong: both methods compute a chain of density-ratios between artificially constructed intermediate distributions, even though TRE defines these intermediate distributions \emph{implicitly}, via samples, whilst AIS typically uses explicit expressions for intermediate densities, and then tries to obtain corresponding samples via MCMC. Moreover, in TRE, we would like to evaluate the learned ratios in \eqref{eq:telescope product} at the \emph{same} input $\x$. In contrast, AIS should only evaluate a ratio $r_k$ at ‘local' samples from e.g.\ $p_k$. Despite these similarities, we would like to emphasise that the connections are only conceptual since AIS and the other methods above are not designed to estimate density-ratios and thus serve fundamentally different purposes.

TRE has conceptual ties with a range of methods in optimisation, statistical physics and machine learning that leverage sequences of intermediate distributions, typically between a complex density $p$ and a simple tractable density $q$. Of particular note are the methods of Simulated Annealing \cite{kirkpatrick1983sim}, Bridge Sampling \& Path Sampling \cite{gelman1998path} and Annealed Importance Sampling (AIS) \cite{neal2001annealed}. Whilst none of these methods estimate density ratios, and thus serve fundamentally different purposes, they leverage similar ideas. In particular, AIS also computes a chain of density-ratios between artificially constructed intermediate distributions. It typically does this by first defining explicit expressions for the intermediate densities, and then trying to obtain samples via MCMC. In contrast, TRE \emph{implicitly} defines the intermediate distributions via samples and then tries to learn the ratios. Additionally, in TRE we would like to evaluate the learned ratios in \eqref{eq:telescope product} at the \emph{same} input $\x$ while AIS should only evaluate a ratio $r_k$ at ‘local' samples from e.g.\ $p_k$.

\subsection{Waymark creation}
\label{sec: methods waymark-creation}
In this paper, we consider two simple, deterministic waymark creation mechanisms: \emph{linear combinations} and \emph{dimension-wise mixing}. We find these mechanisms yield good performance and are computationally cheap. However, we note that other mechanisms are possible, and are a promising topic for future work.

\textbf{Linear combinations.} Given a random pair $\x_0 \sim p_0$ and $\x_m \sim p_m$, define the $k$\textsuperscript{th} waymark via
\begin{align}
\label{eq: linear combination waymarks}
    \x_k &= \sqrt{1 - \alpha_k^2} \ \x_0 + \alpha_k \x_m,  & k = 0, \ldots, m
\end{align}
where the $\alpha_k$ form an increasing sequence from $0$ to $1$, which control the distance of $\x_k$ from $\x_0$. For all of our experiments (except, for illustration purposes, those depicted in Figure \ref{fig:1d single ratio vs tre}), each dimension of $p_0$ and $p_m$ has the same variance\footnote{For MI estimation this always holds, for energy-based modelling this is enforceable via the choice of $p_m$.} and the coefficients in \eqref{eq: linear combination waymarks} are chosen to preserve this variance, with the goal being to match basic properties of the waymarks and thereby make consecutive classification problems harder.

\textbf{Dimension-wise mixing.} An alternative way to `mix' two vectors is to concatenate different subsets of their dimensions. Given a $d$-length vector $\x$, we can partition it into $m$ sub-vectors of length $d/m$, assuming $d$ is divisible by $m$. We denote this as $\x =  (\x[1], \ldots, \x[m])$, where each $\x[i]$ has length $d/m$. Using this notation, define the $k$\textsuperscript{th} waymark via
\begin{align}
\label{eq: dimwise mixing waymarks}
    \x_k &= (\x_m[1], \ \ldots, \ \x_m[k], \ \x_0[k+1], \ \ldots, \ \x_0[m]) & k = 0, \ldots, m
\end{align}
where, again, $\x_0 \sim p_0$ and $\x_m \sim p_m$ are randomly paired.

\textbf{Number and spacing.} Given these two waymark generation mechanisms, we still need to decide the \emph{number} of waymarks, $m$, and, in the case of linear combinations, how the $\alpha_k$ are spaced in the unit interval. We treat these quantities as hyperparameters, and demonstrate in the experiments (Section \ref{sec: experiments}) that tuning them is feasible with a limited search budget.

\subsection{Bridge-building}
\label{sec: methods bridge-building}
Each bridge $r_k(\x ; \thetab_k)$ in \eqref{eq:telescope product} can be learned via binary classification using a logistic loss function as described in Section \ref{sec: dre}. Solving this collection of classification tasks is therefore a multi-task learning (MTL) problem---see \cite{ruder2017overview} for a review. Two key questions in MTL are how to \emph{share parameters} and how to define a \emph{joint objective function}.

\textbf{Parameter sharing}. We break the construction of the bridges $r_k(\x ; \thetab_k)$  into two stages: a (mostly) shared body computing hidden vectors $f_k(\x)$\footnote{For simplicity, we suppress the parameters of $f_k$, and will do the same for $r_k$ in the experiments section.}, followed by bridge-specific heads. The body $f_k$
% \begin{align}
%   & \h_k = f(\x ; \thetab_{\text{sh}}, \s_k, \b_k) \label{eq: shared body f}
% %   & \log r_k(\x; \thetab_k) = g_k(\h_k; \psi_k) + c_k & k \in \{0, \ldots m-1\}, \label{eq: separate heads gk}
% \end{align}
is a deep neural network with shared parameters and pre-activation per-hidden-unit scales and biases for each bridge (see appendix for details). Similar parameter sharing schemes have been successfully used in the multi-task learning literature \cite{de2017condbatchnorm, dumoulin2016artisticstyle}. The heads map the hidden vectors $f_k(\x)$ to the scalar $\log r_k(\x; \thetab_k)$. We use either linear or quadratic mappings depending on the application; the precise parameterisation is stated in each experiment section.

\textbf{TRE loss function}. The TRE loss function is given by the average of the $m$ logistic losses
\begin{align}
 \LT(\thetab)  &=  \frac{1}{m} \sum_{k=0}^{m-1} \L_k(\thetab_k),  \label{eq: tre loss}\\
\L_k(\thetab_k) &= - \E_{\x_k \sim p_k} \log \Big(\frac{r_k(\x_{k}; \thetab_k)}{1 + r_k(\x_{k}; \thetab_k)}  \Big) - 
 \E_{\x_{k+1} \sim p_{k+1}} \log \Big( \frac{1}{1 + r_k(\x_{k+1}; \thetab_k)} \Big). \label{eq: kth tre loss}
\end{align}
This simple \emph{unweighted} average works well empirically. More sophisticated multi-task weighting schemes exist \cite{chen2018gradnorm}, but preliminary experiments suggested they were not worth the extra complexity.

An important aspect of this loss function is that each ratio estimator $r_k$ sees different samples during training. In particular, $r_0$ sees samples close to the real data i.e.\ from $p_0$ and $p_1$, while the final ratio $r_{m-1}$ sees data from $p_{m-1}$ and $p_m$. This creates a potential mismatch between training and deployment, since after learning, we would like to evaluate all ratios at the \emph{same} input $\x$. In our experiments, we do not find this mismatch to be a problem, suggesting that each ratio, despite seeing different inputs during training, is able to generalise to new test points. We speculate that this generalisation is encouraged by parameter sharing, which allows each ratio-estimator to be indirectly influenced by samples from \emph{all} waymark distributions. Nevertheless, we think a deeper analysis of this issue of generalisation deserves further work.

\subsection{TRE applied to mutual information estimation}
The mutual information (MI) between two random variables $\u$ and $\v$ can be written as
\begin{align}
    \label{eq: MI formula}
    I(\u, \v) &= \E_{p(\u, \v)} \Big[ \log r(\u, \v) \Big], & r(\u, \v) = \frac{p(\u, \v)}{p(\u)p(\v)}.
\end{align}
Given samples from the joint density $p(\u, \v)$, one obtains samples from the product-of-marginals $p(\u)p(\v)$ by shuffling the $\v$ vectors across the dataset. This then enables standard density-ratio estimation to be performed.

For TRE, we require waymark samples. To generate these, we take a sample from the joint, $\x_0 = (\u, \v_0)$, and a sample from the product-of-marginals, $\x_m = (\u, \v_m)$, where $\u$ is held fixed and only $\v$ is altered. We then apply a waymark construction mechanism from Section \ref{sec: methods waymark-creation} to generate $\x_k = (\u, \v_k)$, for $k=0, \ldots, m$.

\subsection{TRE applied to energy-based modelling}
\label{sec: tre for EBMs}
% Discriminative density-ratio estimation can be used to learn energy-based models (EBM). All we require is that the denominator density $q$ has an analytically tractable expression, and then $\phi(\x; \thetab) \coloneqq r(\x; \thetab)q(\x)$ is automatically a model of $p$. This model is typically unnormalised---or \emph{energy-based}---because $\phi(\x; \thetab)$ is not constrained to integrate to 1 for all values of $\thetab$.
An energy-based model (EBM) is a flexible parametric family $\{\phi(\x; \thetab)\}$ of non-negative functions, where each function is proportional to a probability-density. Given samples from a data distribution with density $p(\x)$, the goal of energy-based modelling is to find a parameter $\thetab^*$ such that $\phi(\x; \thetab^*)$ is `close' to $c p(\x)$, for some positive constant $c$.

In this paper, we consider EBMs of the form $\phi(\x; \thetab) = r(\x; \thetab) q(\x)$, where $q$ is a known density (e.g. a Gaussian or normalising flow) that we can sample from, and $r$ is an unconstrained positive function. Given this parameterisation, the optimal $r$ simply equals the density-ratio $p(\x)/q(\x)$, and hence the problem of learning an EBM becomes the problem of estimating a density-ratio, which can be solved via TRE. We note that, since TRE actually estimates a product of ratios as stated in Equation \ref{eq:telescope product}, the final EBM will be a product-of-experts model \cite{hinton2006training} of the form $\phi(\x; \thetab) = \prod_{k=0}^{m-1} r_k(\x ; \thetab_k) q(\x)$.

The estimation of EBMs via density-ratio estimation has been studied in multiple prior works, including noise-contrastive estimation (NCE) \cite{Gutmann2012a}, which has many appealing theoretical properties \cite{Gutmann2012a, RiouDurand2018, uehara2018analysis}. Following NCE, we will refer to the known density $q$ as the `noise distribution'.

\vspace{-1ex}
\section{Experiments}
\vspace{-1ex}
\label{sec: experiments}
We include two toy examples illustrating both the correctness of TRE and the fact that it can solve problems which verge on the intractable for standard density ratio estimation. We then demonstrate the utility of TRE on two high-dimensional complex tasks, providing clear evidence that it substantially improves on standard single-ratio baselines.

For experiments with continuous random variables, we use the linear combination waymark mechanisms in \eqref{eq: linear combination waymarks}; otherwise, for discrete variables, we use dimension-wise mixing \eqref{eq: dimwise mixing waymarks}. For the linear combination mechanism, we collapse the $\alpha_k$ into a single spacing hyperparameter, and grid-search over this value, along with the number of waymarks. Details are in the appendix.

\subsection{1d peaked ratio}
\label{sec: 1d peaked gauss experiment}
The basic setup is stated in Figure \ref{fig:1d single ratio subfig}. For TRE, we use quadratic bridges of the form $\log r_k(x) = w_k x^2 + b_k$, where $b_k$ is set to its ground truth value (as derived in appendix), and $w_k$ is reparametrised as $\exp(\theta_k)$ to avoid unnecessary log-scales in Figure \ref{fig:1d single ratio vs tre}. The single ratio-estimation results use the same parameterisation (dropping the subscript $k$).
\begin{wrapfigure}{r}{0.45\linewidth}
\vspace{-0.5cm}
\centering
% left bottom right top
\includegraphics[trim=1ex 0.5ex 0 4.7ex,clip,  width=0.99\linewidth]{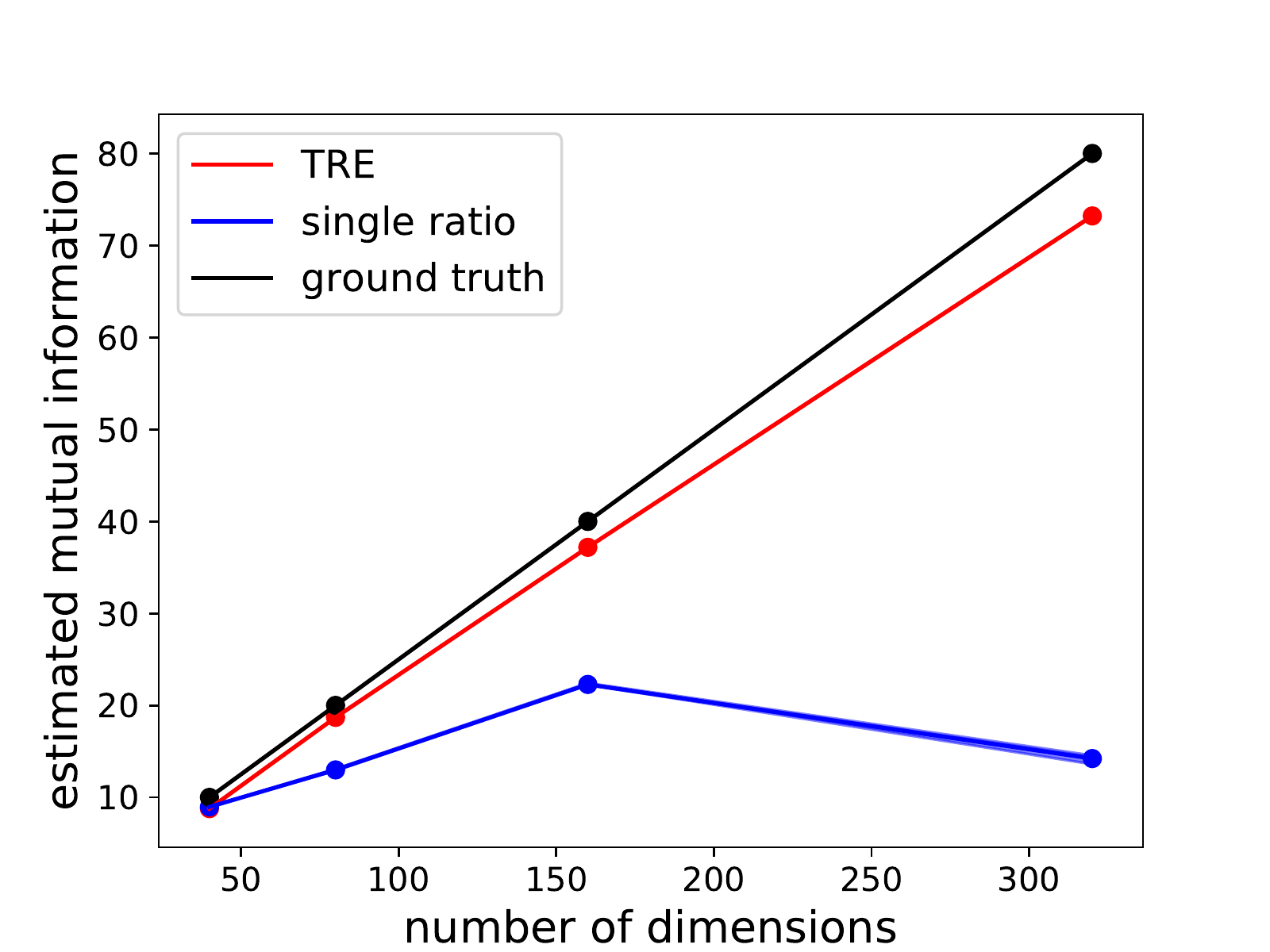}
\caption{High-dimensional Gaussian results, showing estimated MI as a function of the dimensionality. Errors bars were computed over 5 random seeds, but are too small to see.}
\label{fig:correlated gaussians}
\vspace{-1.3cm}
\end{wrapfigure}
Figure \ref{fig:sample efficiency curves} shows the full results. These sample efficiency curves clearly demonstrate that, across all sample sizes, TRE is significantly more accurate than single ratio estimation. In fact, TRE obtains a better solution with 100 samples than single-ratio estimation does with 100,000 samples: a three orders of magnitude improvement.

\subsection{High-dimensional ratio with large MI}
This toy problem has been widely used in the mutual information literature \cite{belghazi2018mutual, poole2019variational}. Let $\x \in \mathbb{R}^{2d}$ be a Gaussian random variable, with block-diagonal covariance matrix, where each block is $2\times2$ with 1 on the diagonal and $0.8$ on the off-diagonal. We then estimate the ratio between this Gaussian and a standard normal distribution. This problem can be viewed as an MI estimation task or an energy-based modelling task---see the appendix for full details.

We apply TRE using quadratic bridges of the form: $\log r_k(\x) = \x^T \W_k \x + b_k$. The results in Figure \ref{fig:correlated gaussians} show that single ratio estimation becomes severely inaccurate for MI values greater than 20 nats. In contrast, TRE can accurately estimate MI values as large as 80 nats for 320 dimensional variables. To our knowledge, TRE is the first discriminative MI estimation method that can scale this gracefully.

\subsection{MI estimation \& representation learning on SpatialMultiOmniglot}
\label{sec: mi experiments}
\begin{figure}
\centering
\begin{subfigure}{0.45\textwidth}
  \centering
  \includegraphics[width=.95\linewidth]{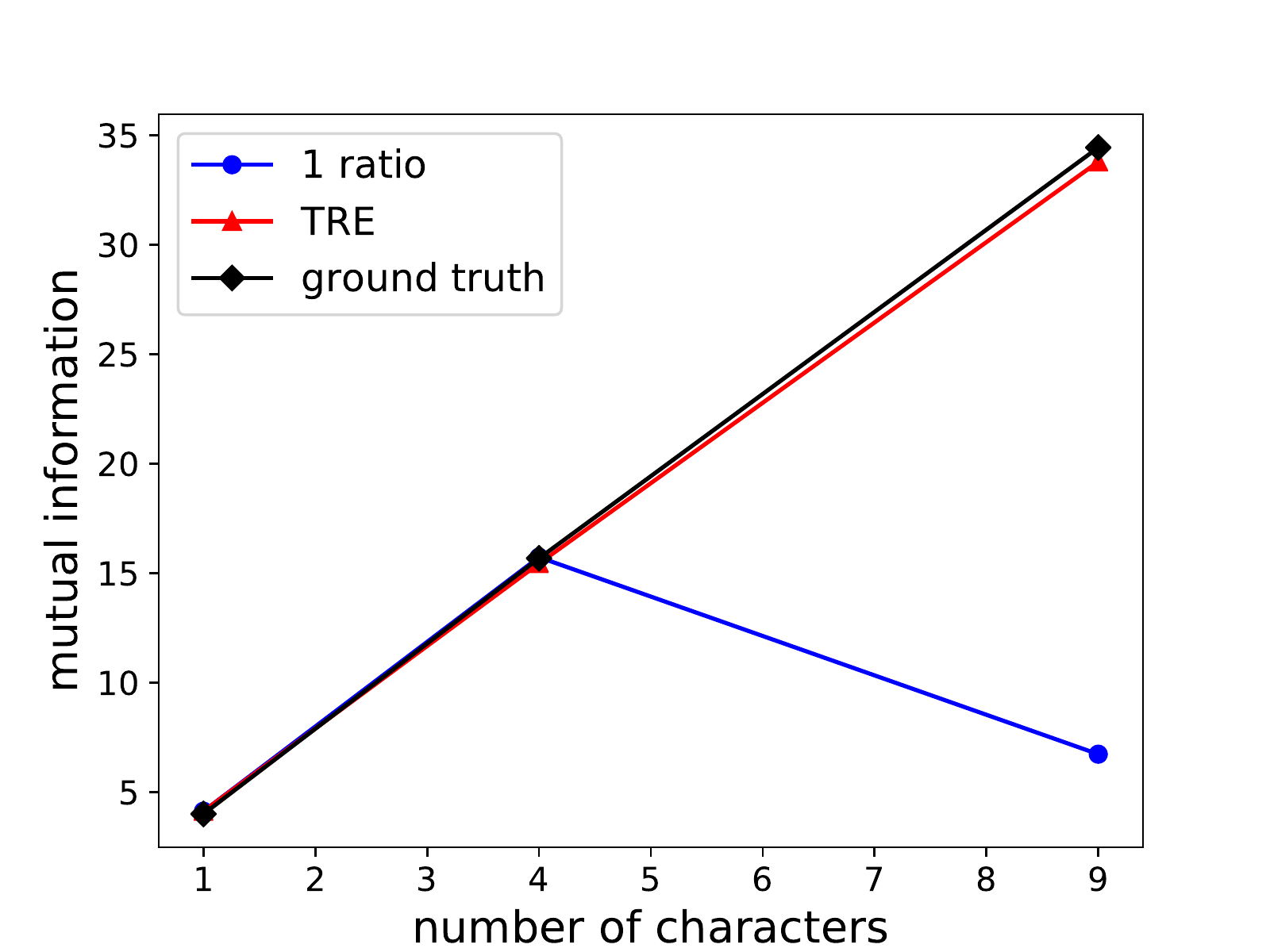}
\end{subfigure}\hfill
\begin{subfigure}{0.45\textwidth}
  \centering
  \includegraphics[width=.95\linewidth]{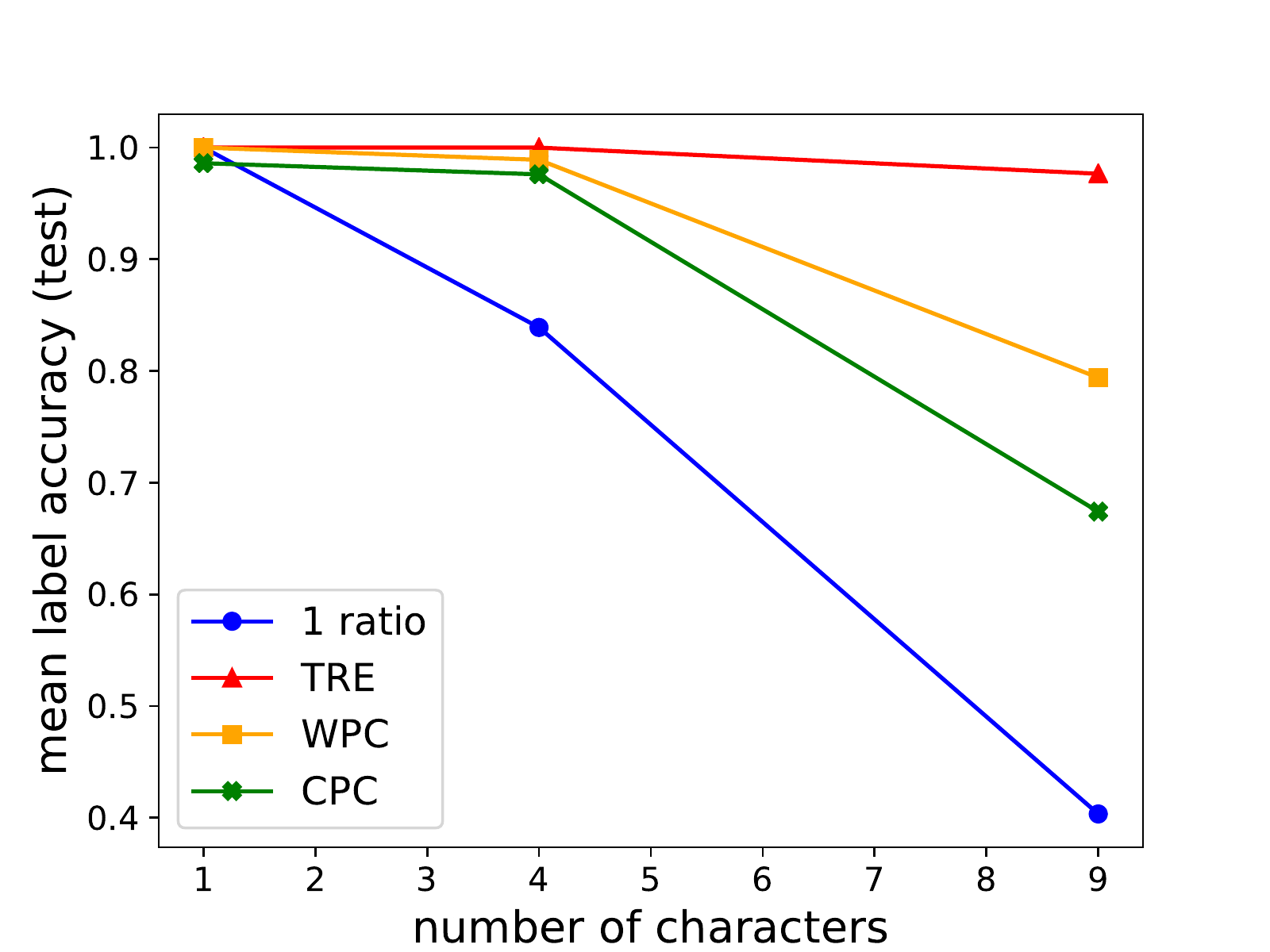}
\end{subfigure}
 \caption{\textbf{Left}: mutual information results. TRE accurately estimates the ground-truth MI even for large values of $\sim 35$ nats. \textbf{Right}: representation learning results. All single density-ratio baselines (this includes CPC \& WPC) degrade significantly in performance as we increase the number of characters from 4 to 9, dropping by 20-60\% in accuracy. In contrast, TRE drops by only $\sim 3\%$.}
  \label{fig: omniglot results}
\end{figure}

We applied TRE to the SpatialMultiOmniglot problem taken from \cite{ozair2019wasserstein}\footnote{We mirror their experimental setup as accurately as possible, however we were unable to obtain their code.} where characters from Omniglot are spatially stacked in an $n \times n$ grid, where each grid position contains characters from a fixed alphabet. Following \cite{ozair2019wasserstein}, the individual pixel values of the characters are not considered random variables; rather, we treat the grid as a collection of $n^2$ categorical random variables whose realisations are the characters from the respective alphabet.
%
%the only source of randomness is the particular subset of characters. This yields discrete random variables with $n^2$ dimensions. 
Pairs of grids, $(\u, \v)$, are then formed such that corresponding grid-positions contain alphabetically consecutive characters. Given this setup, the ground truth MI can be calculated (see appendix).

Each bridge in TRE uses a separable architecture \cite{poole2019variational} given by $\log r_k(\u, \v) = g(\u)^T \W_k f_k(\v)$, where $g$ and $f_k$ are 14-layer convolutional ResNets \cite{he2016deep} and $f_k$ uses the parameter-sharing scheme described in Section \ref{sec: methods bridge-building}. We note that separable architectures are standard in the MI-based representation learning literature \cite{poole2019variational}. We construct waymarks using the dimension-wise mixing mechanism \eqref{eq: dimwise mixing waymarks} with $m=n^2$ (i.e.\ one dimension is mixed at a time).

After learning, we adopt a standard linear evaluation protocol (see e.g.\ \cite{oord2018representation}), where we train \emph{supervised} linear classifiers on top of the output layer $g(\u)$ to predict the alphabetic position of each character in $\u$. We compare our results to those reported in \cite{ozair2019wasserstein}. Specifically, we report their baseline method---contrastive predictive coding (CPC) \cite{oord2018representation}, a state-of-the-art representation learning method based on single density-ratio estimation---along with their variant, Wasserstein predictive coding (WPC).

Figure \ref{fig: omniglot results} shows the results. The left plot shows that only TRE can accurately estimate high MI values of $\sim 35$ nats\footnote{\cite{ozair2019wasserstein} do not provide MI estimates for CPC \& WPC, but \cite{poole2019variational} shows that they are bounded by log batch-size.}. The representation learning results (right) show that all single density-ratio baselines degrade significantly in performance as we increase the number of characters in a grid (and hence increase the MI). In contrast, TRE always obtains greater than $97\%$ accuracy.

\subsection{Energy-based modelling on MNIST}
\label{sec: ebm experiments}
As explained in Section \ref{sec: tre for EBMs}, TRE can be used estimate an energy-based model of the form $\phi(\x; \thetab) = \prod_{k=0}^{m-1} r_k(\x ; \thetab_k) q(\x)$, where $q$ is a pre-specified `noise' distribution from which we can sample, and the product of ratios is given by TRE. In this section, we demonstrate that such an approach can scale to high-dimensional data, by learning energy-based models of the MNIST handwritten digit dataset \cite{lecun1998mnist}. We consider three choices of the noise distribution: a multivariate Gaussian, a Gaussian copula and a rational-quadratic neural spline flow (RQ-NSF) \cite{durkan2019neural} with coupling layers \cite{dinh2017density, kingma2018glow}. Each distribution is first fitted to the data via maximum likelihood estimation---see appendix for details.

Each of these noise distributions can be expressed as an invertible transformation of a standard normal distribution. That is, each random variable has the form $F(\z)$, where $\z \sim \N(0, \I)$. Since $F$ already encodes useful information about the data distribution, it makes sense to leverage this when constructing the waymarks in TRE. Specifically, we can generate linear combination waymarks via \eqref{eq: linear combination waymarks} in $\z$-space, and then map them back to $\x$-space, giving
\begin{align}
\label{eq: z-space linear combination waymarks}
  \x_k &= F(\sqrt{1 - \alpha_k^2} \ F^{-1}(\x_0) + \alpha_k F^{-1}(\x_m)).
\end{align}
% i.e. we linearly interpolate in the $\z$-space of $F$, and then map back to $\x$-space. 
For a Gaussian, $F$ is linear, and hence \eqref{eq: z-space linear combination waymarks} is identical to the original waymark mechanism in \eqref{eq: linear combination waymarks}.

We use the parameter sharing scheme from Section \ref{sec: methods bridge-building} together with quadratic heads. This gives $\log r_k(\x) = -f_k(\x)^T \W_k f_k(\x) - f_k(\x)^T \b_k - c_k$, where we set $f_k$ to be an 18-layer convolutional Resnet and constrain $\W_k$ to be positive definite. This constraint enforces an upper limit on the log-density of the EBM, which has been useful in other work \cite{nash2019autoenergy, ng2011deepenergy}, and improves results here.
\begin{table}
  \caption{Average negative log-likelihood in bits per dimension (bpd, smaller is better). Exact computation is intractable for EBMs, but we provide 3 estimates: Direct/RAISE/AIS. The `Direct' estimate uses the NCE/TRE approximate normalising constant.}
  %The other two estimates are described in the main text.}
  \vspace{3mm}
  \label{table: mnist loglik}
  \centering
  \begin{tabular}{lccc}
    \toprule
    Noise distribution & Noise & Single ratio (NCE) & TRE \\
    \midrule
    && 
    \begin{tabular}{@{}lll}
    \scriptsize{Direct} & \scriptsize{RAISE} & \scriptsize{AIS}
    \end{tabular} & 
    \begin{tabular}{@{}lll}
    \scriptsize{Direct} & \scriptsize{RAISE} & \scriptsize{AIS}
    \end{tabular} \\
    Gaussian  & 2.01 & 
    \begin{tabular}{@{}lll}
    1.96 & 1.99 & 2.01 
    \end{tabular} & 
    \begin{tabular}{@{}lll}
    1.39 & 1.35 & 1.35
    \end{tabular} \\
    Gaussian Copula & 1.40  & 
    \begin{tabular}{@{}lll}
    1.33 & 1.48 & 1.45
    \end{tabular} & 
    \begin{tabular}{@{}lll}
    1.24 & 1.23 & 1.22
    \end{tabular} \\
    RQ-NSF  & 1.12 & 
    \begin{tabular}{@{}lll}
    1.09 & 1.10 & 1.10 
    \end{tabular} & 
    \begin{tabular}{@{}lll}
    1.09 & 1.09 & 1.09
    \end{tabular} \\
    \bottomrule
  \end{tabular}
\end{table}
\begin{figure}
    \begin{subfigure}{0.1\textwidth}
    \vspace{0.2cm}
     \textbf{Gaussian} \vspace{-0.1cm}
     
     \textbf{Copula} \vspace{0.4cm}
     
     \textbf{RQ-NSF}
    \end{subfigure}\hfill
    \begin{subfigure}{0.89\textwidth}
      \hspace{-1.5cm} \textbf{Noise distribution} \hspace{1.5cm} \textbf{Single ratio (NCE)} \hspace{2cm} \textbf{TRE}
      \centering
      \vspace{0.1cm}
      \includegraphics[width=.99\linewidth]{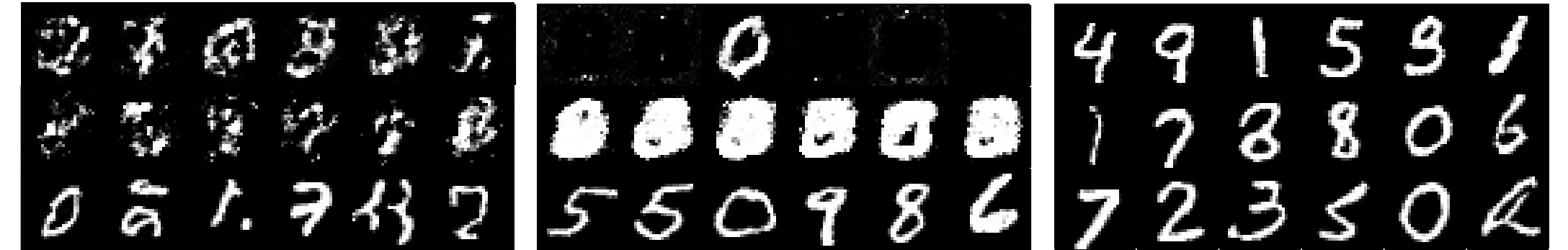}
    \end{subfigure}
  \caption{MNIST samples. Each row pertains to a particular noise distribution. The first block shows exact samples from that distribution. The second \& third blocks show MCMC samples from an EBM learned with NCE \& TRE, respectively.}
  \label{fig:mnist samples}
  \vspace{-0.19cm}
\end{figure}
We evaluate the learned EBMs quantitatively via estimated log-likelihood in Table \ref{table: mnist loglik} and qualitatively via random samples from the model in Figure \ref{fig:mnist samples}. For both of these evaluations, we employ NUTS \cite{hoffman2014no} to perform annealed MCMC sampling as explained in the appendix. This annealing procedure provides two estimators of the log-likelihood: the Annealed Importance Sampling (AIS) estimator \cite{neal2001annealed} and the more conservative Reverse Annealed Importance Sampling Estimator (RAISE) \cite{burda2015accurate}.

The results in Table \ref{table: mnist loglik} and Figure \ref{fig:mnist samples} show that single ratio estimation performs poorly in high-dimensions for simple choices of the noise distribution, and only works well if we use a complex neural density-estimator (RQ-NSF). This illustrates the density-chasm problem explained in Section \ref{sec: dre}. In contrast, TRE yields improvements for all choices of the noise, as measured by the approximate log-likelihood and the visual fidelity of the samples. TRE's improvement over the Gaussian noise distribution is particularly large: the bits per dimension (bpd) is around 0.66 lower, corresponding to an improvement of roughly $360$ nats. Moreover, the samples are significantly more coherent, and appear to be of higher fidelity than the RQ-NSF samples\footnote{We emphasise here that the quality of the RQ-NSF model depends on the exact architecture. A larger model may yield better samples. Thus, we do not claim that TRE generally yields superior results in any sense.}, despite the fact that TRE (with Gaussian noise) has a worse log-likelihood. This final point is not contradictory since log-likelihood and sample quality are known to be only loosely connected \cite{theis2015note}.

Finally, we analysed the sensitivity of our results to the construction of the waymarks and include the results in the appendix. Using TRE with a copula noise distribution as an illustrative case, we found that varying the number of waymarks between 5-30 caused only minor changes in the approximate log-likelihoods, no greater than $0.03$ bpd. We also found that if we omit the $\z$-space waymark mechanism in \eqref{eq: z-space linear combination waymarks}, and work in $\x$-space, then TRE's negative log-likelihood increases to $1.33$ bpd, as measured by RAISE. This is still significantly better than single-ratio estimation, but does show that the quality of the results depends on the exact waymark mechanism.

\section{Conclusion}
We introduced a new framework---Telescoping density-Ratio Estimation (TRE)---for learning density-ratios that, unlike existing discriminative methods, can accurately estimate ratios between extremely different densities in high-dimensions.

TRE admits many exciting directions for future work. Firstly, we would like a deeper theoretical understanding of why it is so much more sample-efficient than standard density-ratio estimation. The relationship between TRE and standard methods is structurally similar to the relationship between annealed importance sampling and standard importance sampling. Thus, exploring this connection further may be fruitful. Relatedly, we believe that TRE would benefit from further research on waymark mechanisms. We presented simple mechanisms that have clear utility for both discrete and continuous-valued data. However, we suspect more sophisticated choices may yield improvements, especially if one can leverage domain or task-specific assumptions to intelligently decompose the density-ratio problem. Lastly, whilst this paper has focused on the logistic loss, it would be interesting to more deeply investigate TRE with other discriminative loss functions.

% other discriminative losses can be used for TRE, and investigating this more deeply would be interesting. 
% it would be interesting to more deeply investigate other discriminative loss functions for TRE.
% it would be interesting to more deeply investigate TRE with other discriminative loss functions.

\section*{Broader Impact}
As outlined in the introduction, density-ratio estimation is a foundational tool in machine learning with diverse applications. Our work, which improves density-ratio estimation, may therefore increase the scope and power of a wide spectrum of techniques used both in research and real-world settings. The broad utility of our contribution makes it challenging to concretely assess the societal impact of the work. However, we do discuss here two applications of density-ratio estimation with obvious potential for positive \& negative impacts on society.

Generative Adversarial Networks \cite{goodfellow2014generative} are a popular class of models which are often trained via density-ratio estimation and are able to generate photo-realistic image/video content. To the extent that TRE can enhance GAN training (a topic we do not treat in this paper), our work could conceivably lead to enhanced `deepfakes', which can be maliciously used in fake-news or identity fraud.

More positively, density-ratio estimation is being used to correct for dataset bias, including the presence of skewed demographic factors like race and gender \cite{grover2019fair}. While we are excited about such applications, we emphasise that density-ratio based methods are not a panacea; it is entirely possible for the technique to introduce new biases when correcting for existing ones. Future work should continue to be mindful of such a possibility, and look for ways to address the issue if it arises.

\begin{ack}
Benjamin Rhodes was supported in part by the EPSRC Centre for Doctoral Training in Data Science, funded by the UK Engineering and Physical Sciences Research Council (grant EP/L016427/1) and the University of Edinburgh. Kai was supported by Edinburgh Huawei Research Lab in the University of Edinburgh, funded by Huawei Technologies Co. Ltd.
\end{ack}

\small
\bibliography{tre_bib}

\clearpage
\begin{appendix}
\section{ResNet architectures with parameter sharing}

\begin{figure}[h]
  \label{table: resnet}
    \begin{subfigure}{0.3\textwidth}
    \centering
    \begin{tabular}{c}
    \toprule
      $5 \times 5$ conv with \\ $3 \times 3$ strides, 32n \\
      \midrule
      CondionalScaleShift \\
      \midrule
      CondResBlock down, 32n \\
      \midrule
      CondResBlock, 32n \\
      \midrule
      CondResBlock down, 64n \\
      \midrule
      CondResBlock, 64n \\
      \midrule
      CondResBlock down, 64n \\
      \midrule
      CondResBlock, 64n \\
      \midrule
      GlobalSumPooling \\
      \midrule
      Dense, 300n \\
    \bottomrule
  \end{tabular}
  \caption{SpatialMultiOmniglot architecture. The multiplier $n$ refers to the width/height of a datapoint, which is an $n \times n$ grid.}
  \label{table: multiomniglot architecture}
  \end{subfigure}\hfill
  \begin{subfigure}{0.3\textwidth}
  \centering
  \begin{tabular}{c}
 
    \toprule
      $3 \times 3$ conv, 64 \\
      \midrule
      CondionalScaleShift \\
      \midrule
      CondResBlock down, 64 \\
      \midrule
      AttentionBlock \\
      \midrule
      CondResBlock, 64 \\
      \midrule
      CondResBlock down, 64 \\
      \midrule
      CondResBlock, 64 \\
      \midrule
      CondResBlock down, 128 \\
      \midrule
      CondResBlock, 128 \\
      \midrule
      CondResBlock down, 128 \\
      \midrule
      CondResBlock, 128 \\
      \midrule
      GlobalSumPooling \\
       \midrule
       Dense, 128 \\
    \bottomrule
  \end{tabular}
  \caption{MNIST architecture.}
  \label{table: mnist architecture}
  \end{subfigure}\hfill
  \begin{subfigure}{0.32\textwidth}
    \centering
    \includegraphics[width=0.99\textwidth]{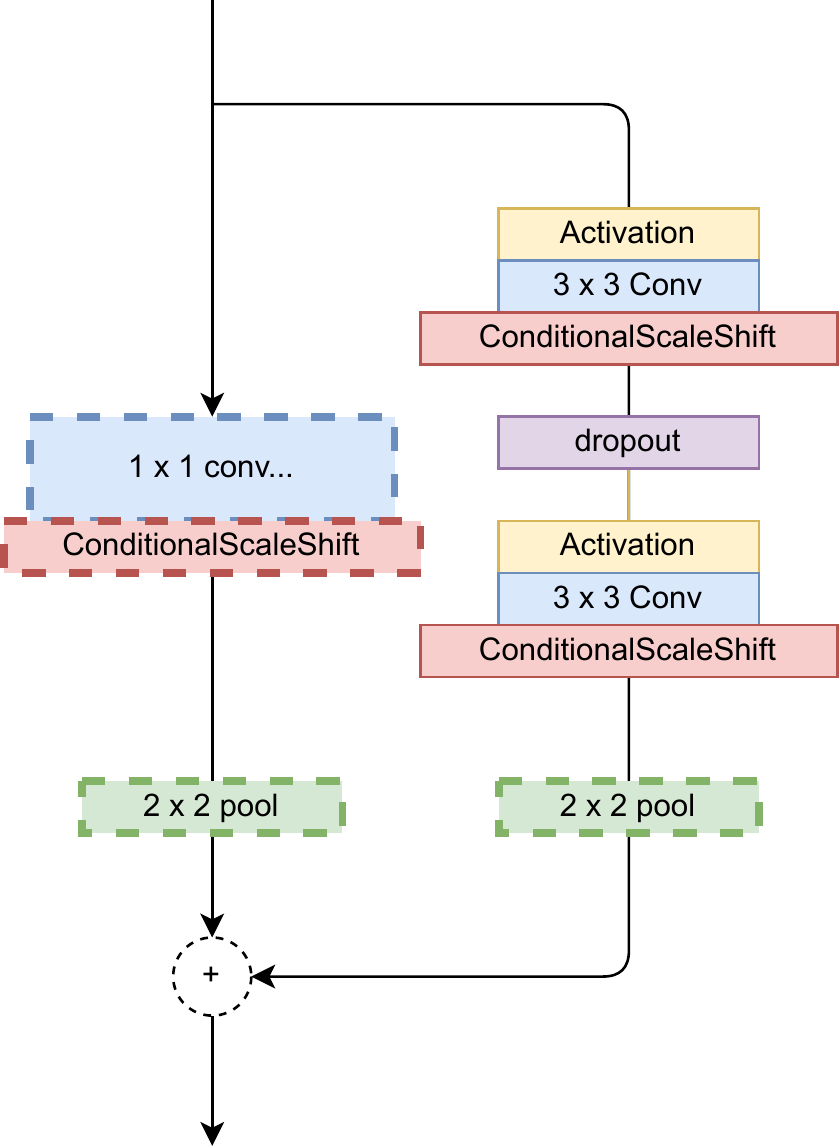}
    \caption{CondResBlock. Dashed boxes denote layers that are not always present. The $1 \times 1$ conv layer (and the associated CondScaleShift) is only used in blocks where the channel size is altered. The $2 \times 2$ pool layer is only used for `down' blocks.}
    \label{fig: condresblock}
  \end{subfigure}
  \caption{Convolutional ResNet architectures}
  \label{fig: architectures}
\end{figure}

In Figure \ref{fig: architectures}, we give the exact architectures for the $f_k$ used in the two high-dimensional experiments on SpatialMultiOmniglot and MNIST. These $f_k$ output a hidden vector for the $k$\textsuperscript{th} bridge, which is then mapped to the scalar value of the log-ratio, as stated in each experiment section. All convolution operations share their parameters across the bridges, and are thus independent of $k$.

The only difference between our conditional residual blocks (i.e.\ `CondResBlocks') and a standard residual block is the use of `ConditionalScaleShift' layers. These layers map a hidden vector $\z_k$ to a hidden vector of the same size, $\z'_k$, via
\begin{align}
    \z'_k = \s_k  \odot \z_k + \b_k
\end{align}
where $\s_k$ and $\b_k$ are bridge-specific parameters and $\odot$ denotes element-wise multiplication. This operation could be thought of as class-conditional Batch Normalisation (BN) \cite{de2017condbatchnorm} without the normalisation. We did not investigate the use of BN, since many energy-based modelling papers (e.g.\ \cite{du2019implicit}) found it to harm performance. We did perform preliminary experiments with Instance Normalisation \cite{Ulyanov2017ImprovedTN} in the context of energy-based modelling, finding it to be harmful to performance.

For the MNIST energy-based modelling experiments, we use average pooling operations since other work \cite{song2019generative, du2019implicit} has found this to produce higher quality samples than max pooling. For the SpatialMultiOmniglot experiments, we grid-search over average pooling and max pooling. For both sets of experiments, we use LeakyRelu activations with a slope of $0.3$.

The MNIST architecture includes an attention block \cite{zhang2019attention} which has been used in GANs to model long-range dependencies in the input image. We found that this attention layer did not yield improvements in estimated log-likelihood, but we think it \emph{may} yield slightly more globally coherent samples. We note that that another commonly used feature in recent GAN and EBM architectures is Spectral Normalisation (SN) \cite{miyato2018spectral}. Our preliminary experiments suggested that SN was not beneficial for performance. That said, all of our negative results should be taken with a grain of salt, given the preliminary nature of the experiments.

\section{Waymark number and spacing}
\label{sec: wmark spacing}
\begin{table}[b]
\begin{threeparttable}
  \caption{Waymark hyperparameters for each experiment. Curly braces $\{ \}$ denote grid-searches.}
  \label{table: wmark spacing}
    \vspace{3mm}
  \centering
  \begin{tabular}{lllll} %{lcccc}
   \\
    \toprule 
    experiment & mechanism & $m$ & spacing & $p$ \\
    \midrule
    1d peaked ratio & linear combo & 4 & Eq.\ \ref{eq: increasing alpha_k} & $\{ 1, 2, \ldots, 7, 8 \}$ \\
    high dim, high MI & linear combo & $\frac{d}{40} \times \{1, 2, 3, 4\}$ & Eq.\ \ref{eq: increasing alpha_k} & 1 \\
    SpatialMultiOmniGlot & dim-wise mix & $d$ & N/A & N/A \\
    MNIST (z-space) & linear combo & $\{5, 10, 15, 20, 25, 30\}$ & Eq.\ \ref{eq: increasing alpha_k} & 1 \\
    MNIST (x-space) & linear combo & $\{5, 10, 15, 20, 25, 30\}$ & Eq.\ $\{$ \ref{eq: increasing alpha_k}, \ref{eq: symmetric alpha_k} $\}$ & $\{ 1, 1.5, 2\}$ \\
    \bottomrule
  \end{tabular}
  \begin{tablenotes}
  \small
      \item Note: $d$ refers to the dimensionality of the dataset, which is varied for certain experiments.
  \end{tablenotes}
  \end{threeparttable}
\end{table}
As stated in the main text, the number and (in the case of linear combinations) the spacing of the waymarks are treated as hyperparameters. Finding good values of these hyperparameters is made simpler by the following observations.
\begin{itemize}
    \item If any of the TRE logistic losses saturate close to 0 during learning, then this indicates that the density-chasm problem has occured for that bridge, and we can terminate the run. 
    \item As illustrated by our sensitivity analysis for MNIST (see Figure \ref{fig: mnist wmark sensitivity}) it seems that, past a certain point, performance plateaus with the addition of extra waymarks. The fact that it plateaus, and does not decrease, is good news since it indicates that there is little risk of `overshooting', and obtaining a bad model by having too many waymarks.
\end{itemize}

We now recall the linear combinations waymark mechanism, given by
\begin{align}
\label{eq: appendix linear combination waymarks}
    \x_k &= \sqrt{1 - \alpha_k^2} \ \x_0 + \alpha_k \x_m,  & k = 0, \ldots, m.
\end{align}
where $m$ is the number of waymarks. We consider two ways of reducing the coefficients $\alpha_k$ to a function of a single spacing hyperparameter $p$ via
\begin{align}
    \alpha_k &= (k/m)^p, & k = 0, \ldots, m, \label{eq: increasing alpha_k}\\
    \alpha_k &= 
\left\{\begin{array}{lr}
        (k/m)^p, & \text{for } k\leq m/2 \\
        1 - ((m-k)/m)^p, & \text{for } k\geq m/2\\
        \end{array}\right\} & k = 0, \ldots, m. \label{eq: symmetric alpha_k}
\end{align}
Both mechanisms yield linearly spaced $\alpha_k$ when $p=1$. For the first mechanism in \eqref{eq: increasing alpha_k}, setting $p > 1$ means the gaps between waymarks \emph{increase} with $k$ (and conversely decrease if $p < 1$). The spacing mechanism in \eqref{eq: symmetric alpha_k} is a kind of symmetrised version of \eqref{eq: increasing alpha_k}.

Table \ref{table: wmark spacing} shows the grid-searches we performed for all experiments. We note that these weren't always all performed in parallel. When using linear combinations, we typically set $p=1$ initially and searched over values of $m$. If, for all values of $m$ tested, one of the TRE logistic losses saturated close to 0, then we would expand our search space and test different values of $p$.
%\vspace{-0.5cm}

\section{Minibatching}
\label{sec: appendix minibatching}
Recall that the TRE loss is a sum of logistic losses:
\begin{align}
 \LT(\thetab)  &=  \frac{1}{m} \sum_{k=0}^{m-1} \L_k(\thetab_k),  \label{eq: tre loss}\\
\L_k(\thetab_k) &= - \E_{\x_k \sim p_k} \log \Big(\frac{r_k(\x_{k}; \thetab_k)}{1 + r_k(\x_{k}; \thetab_k)}  \Big) - 
 \E_{\x_{k+1} \sim p_{k+1}} \log \Big( \frac{1}{1 + r_k(\x_{k+1}; \thetab_k)} \Big). \label{eq: kth tre loss}
\end{align}
When generating minibatch estimates of this loss, we can either sample from each $p_k$ independently, or we can \emph{couple} the samples. By ‘couple', we mean first drawing $B$ samples each from $p_0$ and $p_m$, randomly pairing members from each set, and then, for each pair, constructing all possible intermediate waymark samples to obtain a final minibatch of size $B \times M$. Coupling in this way means that the gradient of \eqref{eq: tre loss} w.r.t. $\thetab$ is estimated using shared sources of randomness, which can act as a form of variance reduction \citep{mohamed2019monte}. 

In all of our experiments, we use coupling when forming minibatches, since we found it to be useful in some preliminary investigations. However, coupling does have memory costs: the number of independent samples drawn from the data distribution, $B$, may need to be very small for the full minibatch, $B \times M$, to fit into memory. We speculate that as $B$ becomes sufficiently small, coupled minibatches will produce inferior results to non-coupled minibatches (which can use a greater number of independent real data samples). Empirical investigation of this claim is left to future work.

\section{1d peaked ratio toy experiment}
In this experiment we estimate the ratio $p_0/p_m$, where both densities are Gaussian, $p_0 = \N(0, \sigma_0^2)$ and $p_m = \N(0, \sigma_m^2)$, where $\sigma_0 = 10^{-6}$ and $\sigma_m = 1$. We generate waymarks using the linear combinations mechanism \eqref{eq: appendix linear combination waymarks}, which implies that each waymark distribution is Gaussian, since linear combinations of Gaussian random variables are also Gaussian. Specifically, the waymark distributions have the form
\begin{align}
    p_k(x) &= \mathcal{N}(x; 0, \sigma_k^2),  & \text{where} \ \ \sigma_k = \big[ (1-\alpha_k^2) \sigma_0^2 + \alpha_k^2 \sigma_m^2 \big]^{\frac{1}{2}}.
\end{align}
where the $\sigma_k$ form an increasing sequence between $\sigma_0$ and $\sigma_m$. The log-ratio between two waymark distributions is therefore given by
\begin{align}
\label{eq: 1d gauss true waymark ratio}
    \log \frac{p_k(x)}{p_{k+1}(x)} = \log \big( \frac{\sigma_{k+1}}{\sigma_{k}} \big) + \big( \frac{1}{2 \sigma_{k+1}^2} - \frac{1}{2\sigma_k^2} \big) x^2.
\end{align}
We parameterise the bridges in TRE as
\begin{align}
\label{eq: 1d bridge parameterisation}
    \log r_k(x; \theta_k) = \log \big( \frac{\sigma_{k+1}}{\sigma_{k}} \big) - \exp(\theta_k) x^2,
\end{align}
where the quadratic coefficient $-\exp(\theta_k)$ is always negative. We note that this model is well-specified since it contains the ground-truth solution in \eqref{eq: 1d gauss true waymark ratio}.

The bridges can then be combined via summation to provide an estimate of the original log-ratio
\begin{align}
    \log \frac{p_0(x)}{p_m(x)} &\approx \sum_{k=0}^{m-1} \log r_k(x; \theta_k) \\
    &= \log \big( \frac{\sigma_m}{\sigma_{0}} \big) - \sum_{k=0}^{m-1} \exp(\theta_k) x^2 \\
    &= \log \big( \frac{\sigma_m}{\sigma_{0}} \big) - \exp(\theta_{TRE}) x^2 \label{eq: 1d theta tre}
\end{align}
Where $\theta_{TRE} = \log ( \sum_{k=0}^{m-1} \exp(\theta_k))$. We observe that \eqref{eq: 1d theta tre} has the same form as \eqref{eq: 1d bridge parameterisation} if we were to set $m=1$ in \eqref{eq: 1d bridge parameterisation} (i.e.\ if we use a single bridge). Hence $\theta_{TRE}$ can be directly compared to the parameter value we would obtain if we used single density-ratio estimation. This is precisely the comparison we make in Figure 1a and Figure 2 of the main text.

\subsection{The density chasm problem for non-logistic loss functions}

In the main paper, we illustrated the density-chasm problem for the logistic loss using the 1d peaked ratio experiment. Here, we illustrate precisely the same phenomenon for the NWJ/MINE-f loss \cite{nguyen2010estimating, belghazi2018mutual} and a Least Squares (LSQ) loss used by \cite{mao2017least}. The loss functions are given by
\begin{align}
    \LNWJ(\thetab) &= -\E_p \left[ \log r(\x; \thetab) \right] - 1 + \E_q \left[ r(\x; \thetab) \right] \\
    \LLSQ(\thetab) &= \frac{1}{2} \E_p \left[ (\sigma(\log(r(\x; \thetab))) - 1)^2 \right] + \frac{1}{2} \E_q \left[ (\sigma(\log(r(\x; \thetab))))^2\right], \label{eq: lsq loss}
\end{align}
where the $\sigma$ in \eqref{eq: lsq loss} denotes the sigmoid function.

\begin{figure}
\centering
 \large \textbf{NWJ loss}
\begin{subfigure}{0.95\textwidth}
  \centering
  \includegraphics[width=.99\linewidth]{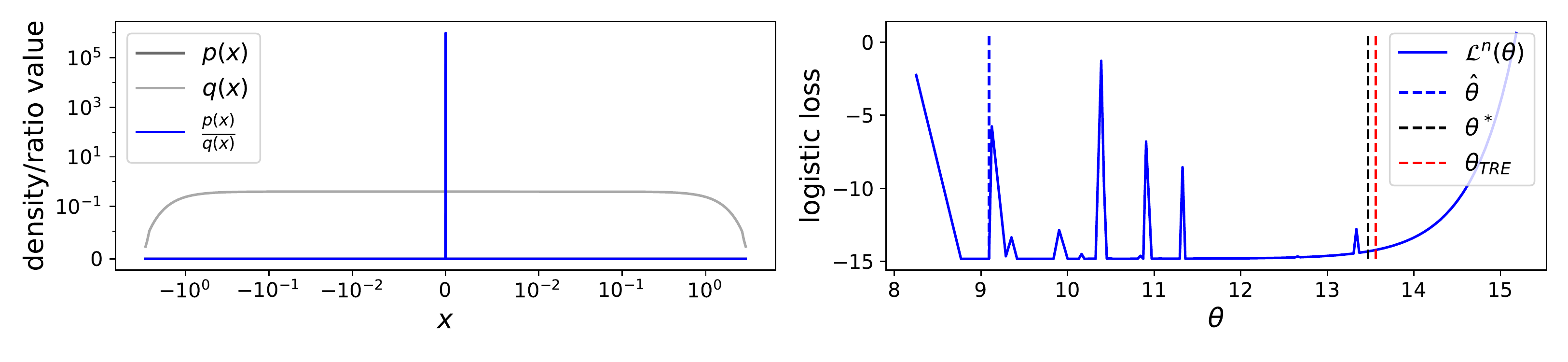}
  \label{fig:1d single ratio nwj subfig}
\end{subfigure}
\begin{subfigure}{0.95\textwidth}
  \centering
    \hspace{-0.4cm}
        \begin{tabular}{c c c c}
        \hspace{-12mm}
        \Large $\boldsymbol{\frac{p}{q}}$
        \hspace{4mm} \Large $\boldsymbol{=}$
        \hspace{3mm} \Large $\boldsymbol{\frac{p}{p_1}}$ &
        \hspace{-20mm} \Large $\boldsymbol{\times}$
        \hspace{12mm} \Large $\boldsymbol{\frac{p_1}{p_2}}$ &
        \hspace{-20mm} \Large $\boldsymbol{\times}$
        \hspace{12mm} \Large $\boldsymbol{\frac{p_2}{p_3}}$ &
        \hspace{-20mm} \Large $\boldsymbol{\times}$
        \hspace{12mm} \Large $\boldsymbol{\frac{p_3}{q}}$ \\
        \includegraphics[width=.24\linewidth]{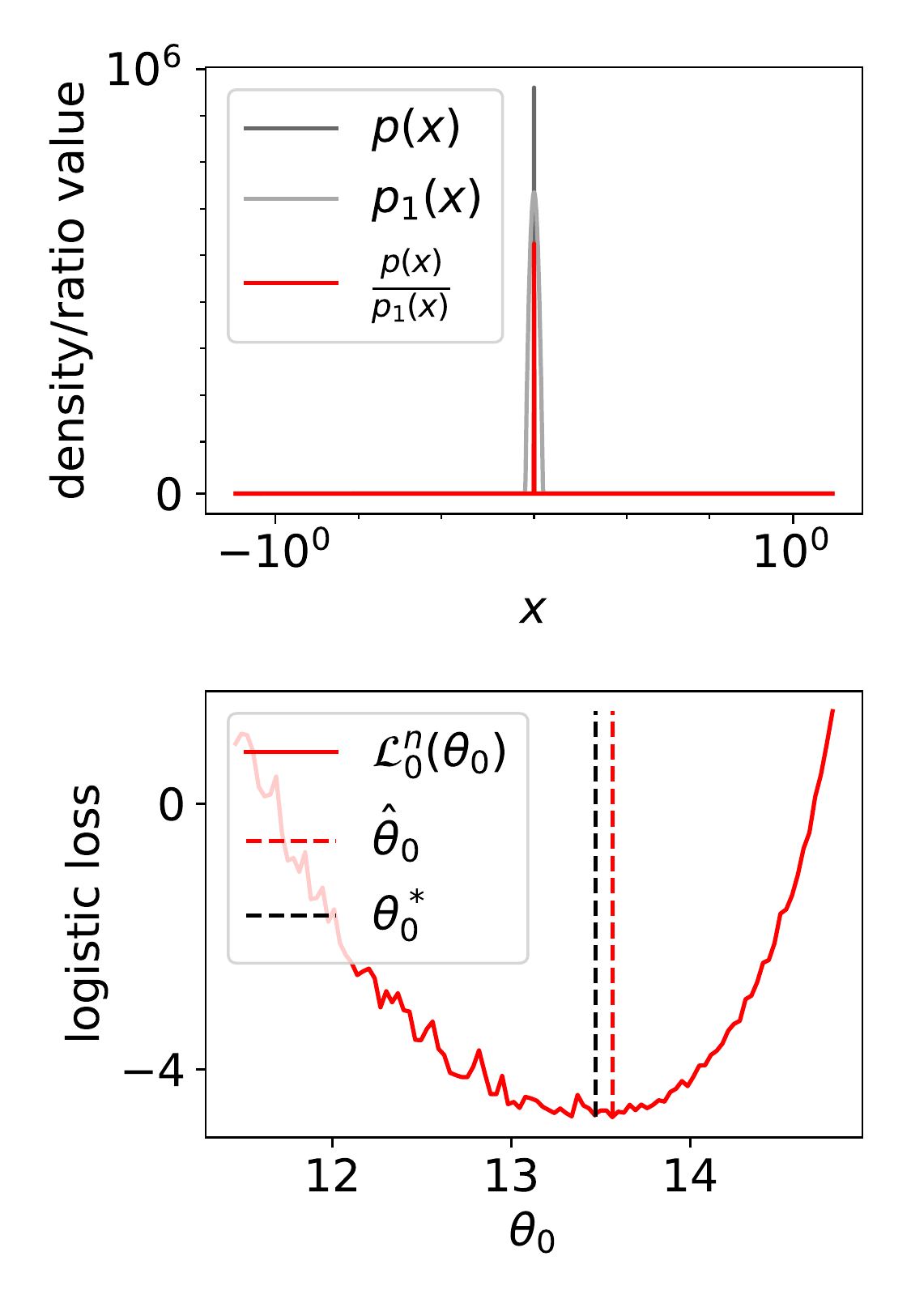} &
        \hspace{-0.5cm} \includegraphics[width=.24\linewidth]{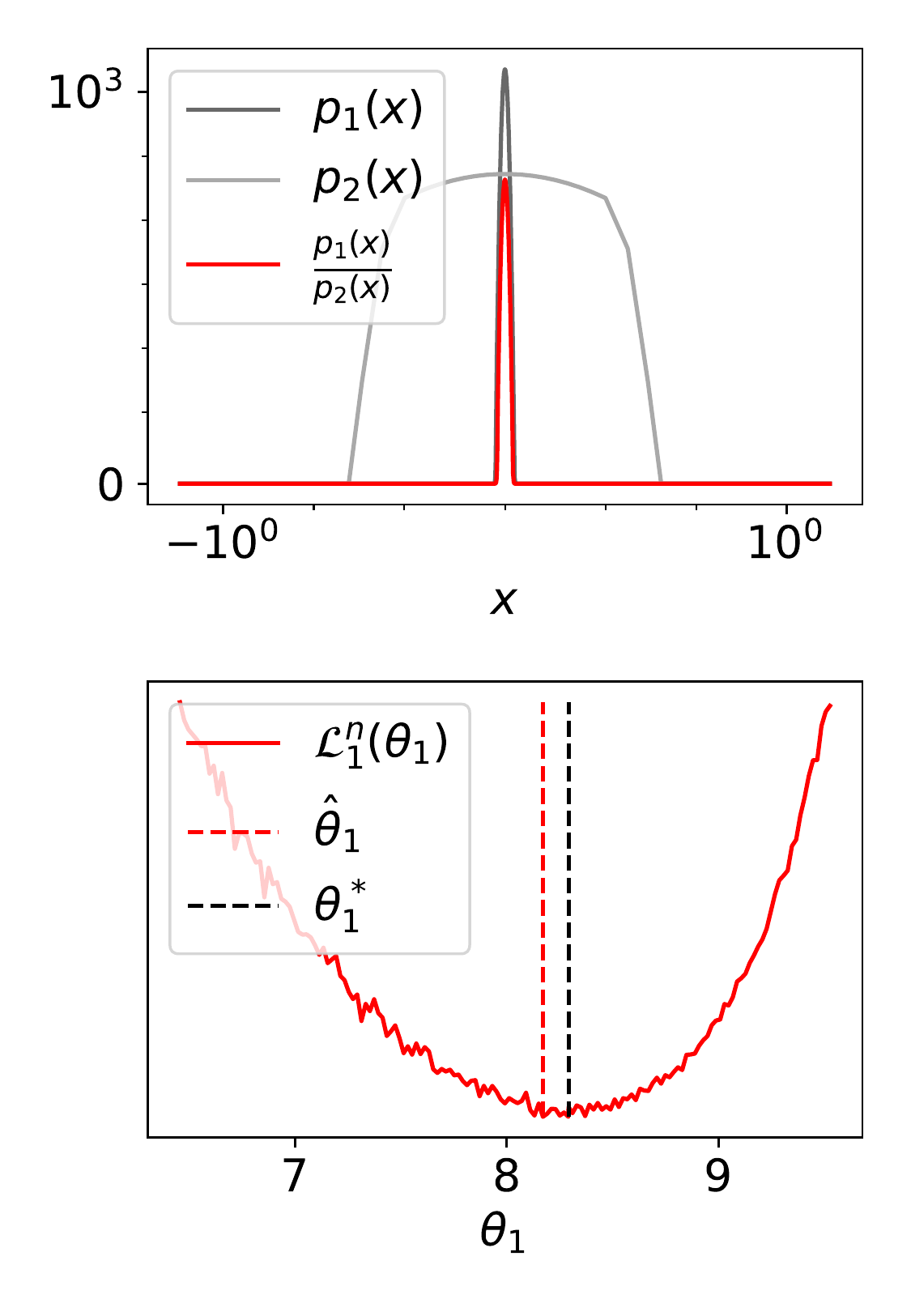} &
        \hspace{-0.5cm}
        \includegraphics[width=.24\linewidth]{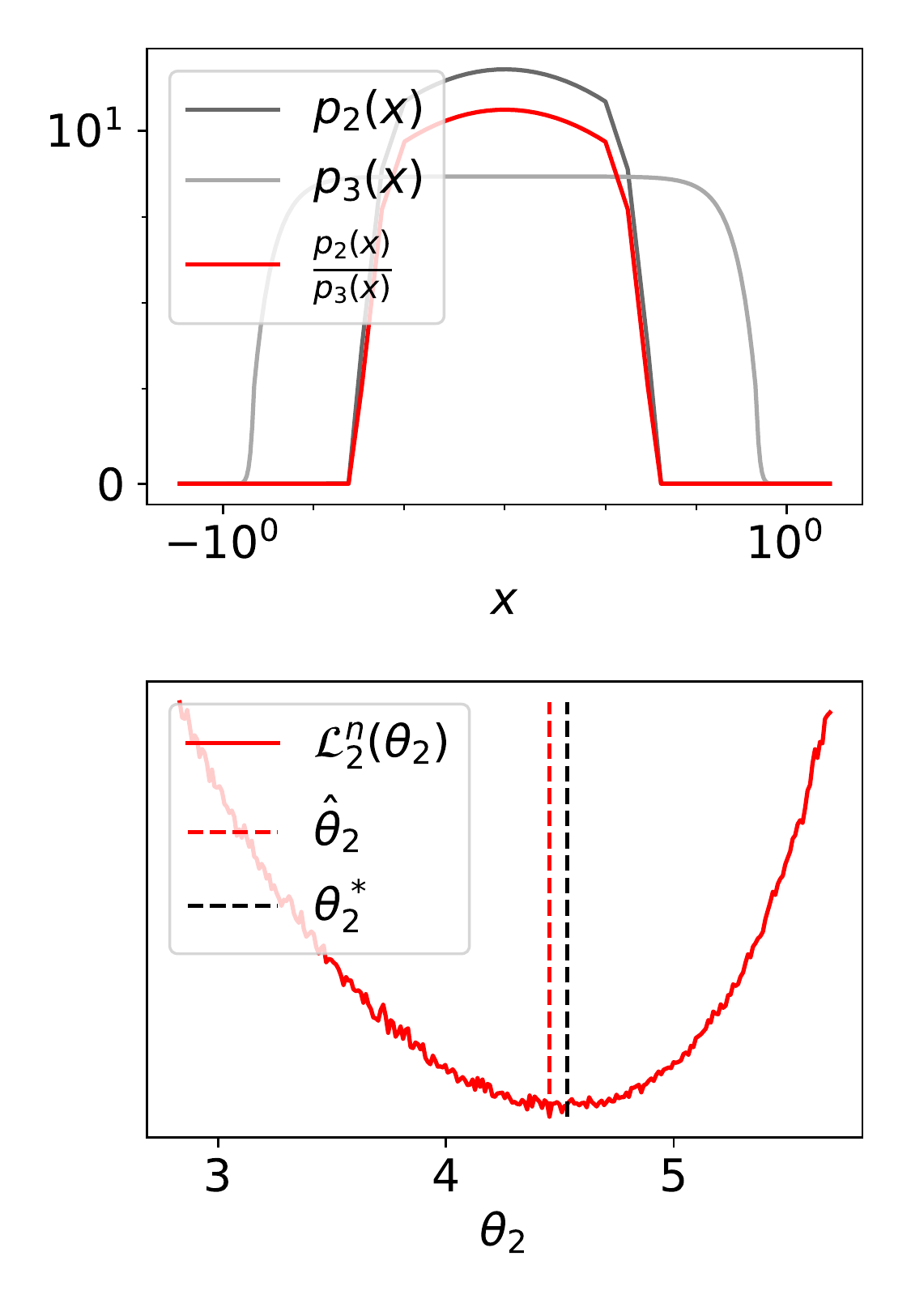} &
        \hspace{-0.5cm}
        \includegraphics[width=.24\linewidth]{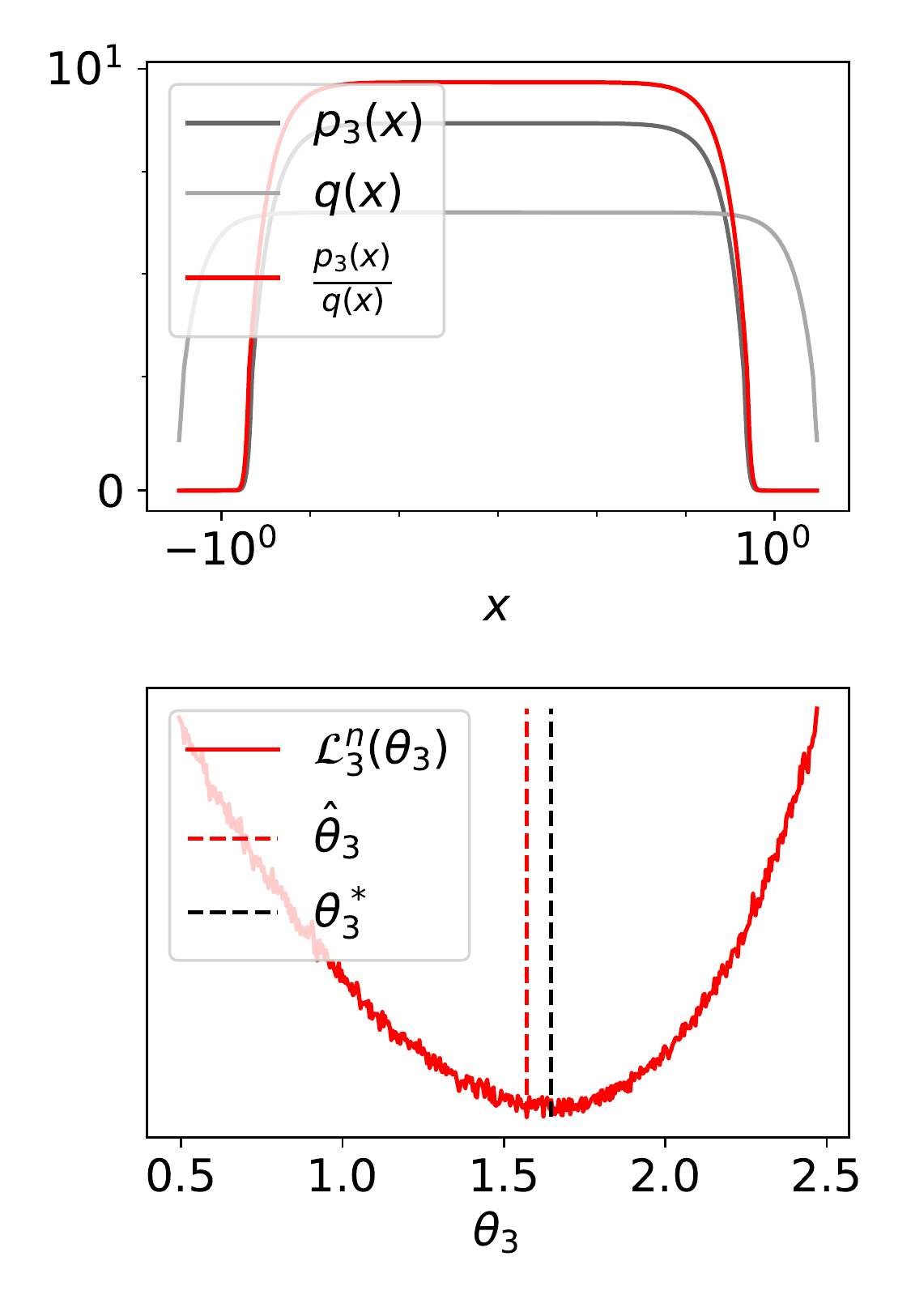}
    \end{tabular}
\label{fig:1d tre nwj subfig}
\end{subfigure}
\caption{Replica of Figure 1 from the main text, except that we use the NWJ/MINE-f loss \cite{nguyen2010estimating, belghazi2018mutual} for both the single ratio estimator \& for each ratio in TRE.}
\label{fig:1d single ratio vs tre for nwj}
\vspace{1cm}
\large \textbf{Least-square loss}
\centering
\begin{subfigure}{0.95\textwidth}
  \centering
  \includegraphics[width=.99\linewidth]{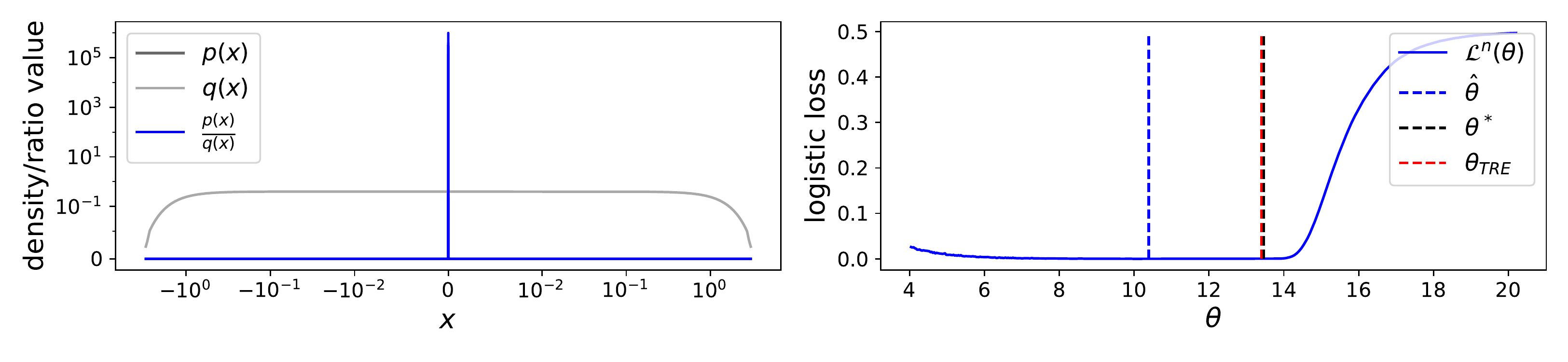}
  \label{fig:1d single ratio lsq subfig}
\end{subfigure}
\begin{subfigure}{0.95\textwidth}
  \centering
    \hspace{-0.4cm}
        \begin{tabular}{c c c c}
        \hspace{-12mm}
        \Large $\boldsymbol{\frac{p}{q}}$
        \hspace{4mm} \Large $\boldsymbol{=}$
        \hspace{3mm} \Large $\boldsymbol{\frac{p}{p_1}}$ &
        \hspace{-20mm} \Large $\boldsymbol{\times}$
        \hspace{12mm} \Large $\boldsymbol{\frac{p_1}{p_2}}$ &
        \hspace{-20mm} \Large $\boldsymbol{\times}$
        \hspace{12mm} \Large $\boldsymbol{\frac{p_2}{p_3}}$ &
        \hspace{-20mm} \Large $\boldsymbol{\times}$           
        \hspace{12mm} \Large $\boldsymbol{\frac{p_3}{q}}$ \\
        \includegraphics[width=.24\linewidth]{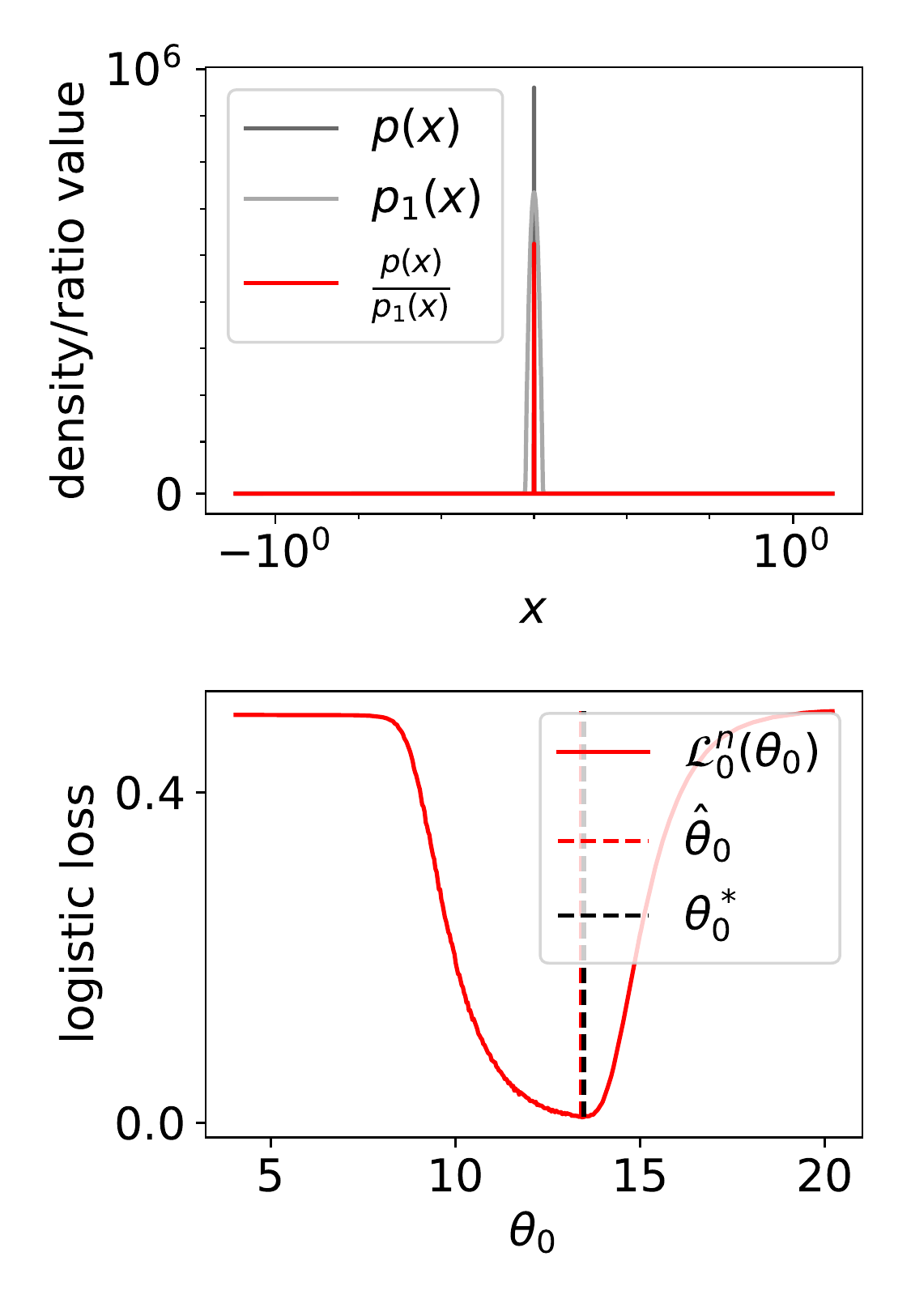} &
        \hspace{-0.5cm} \includegraphics[width=.24\linewidth]{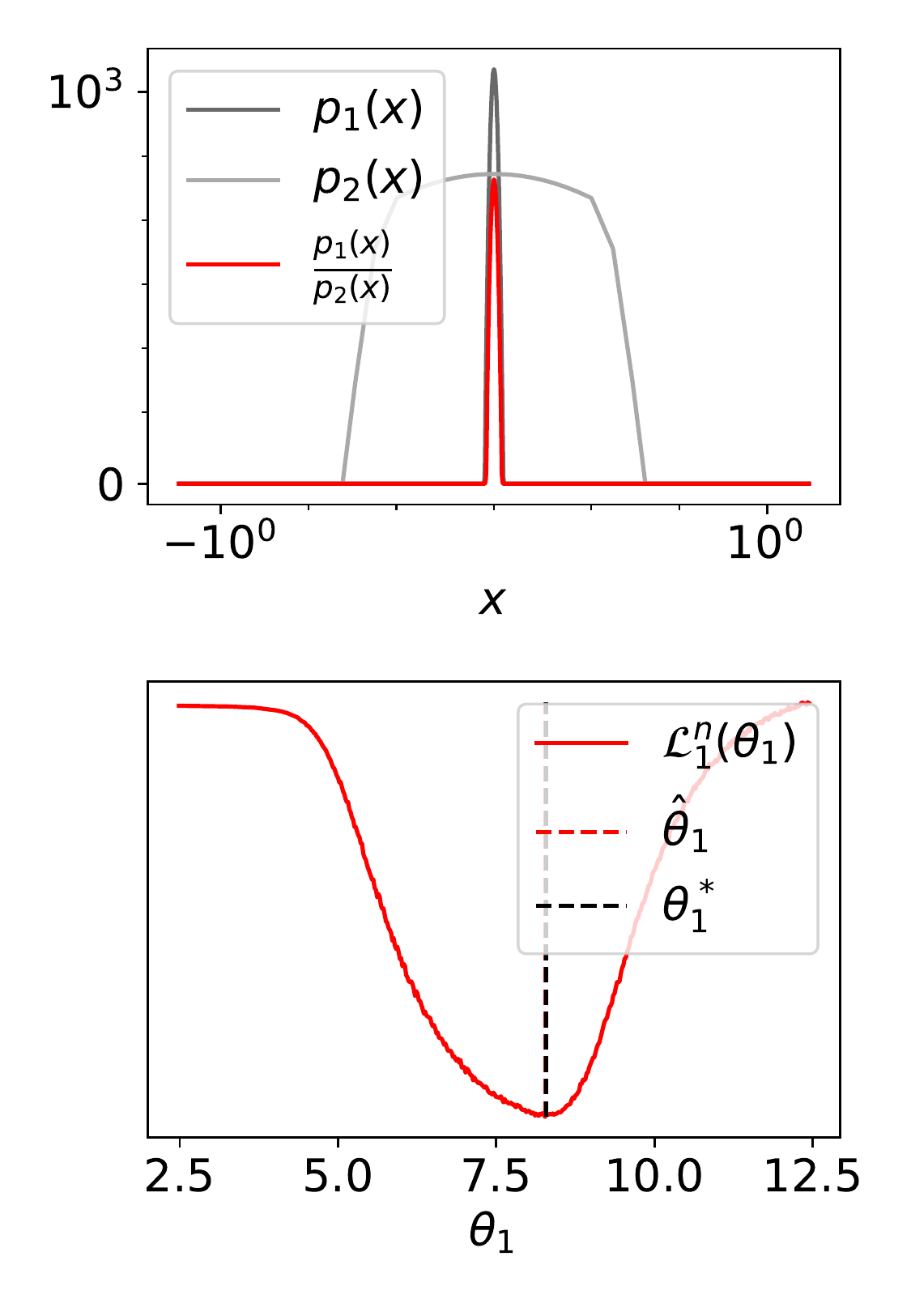} &
        \hspace{-0.5cm}
        \includegraphics[width=.24\linewidth]{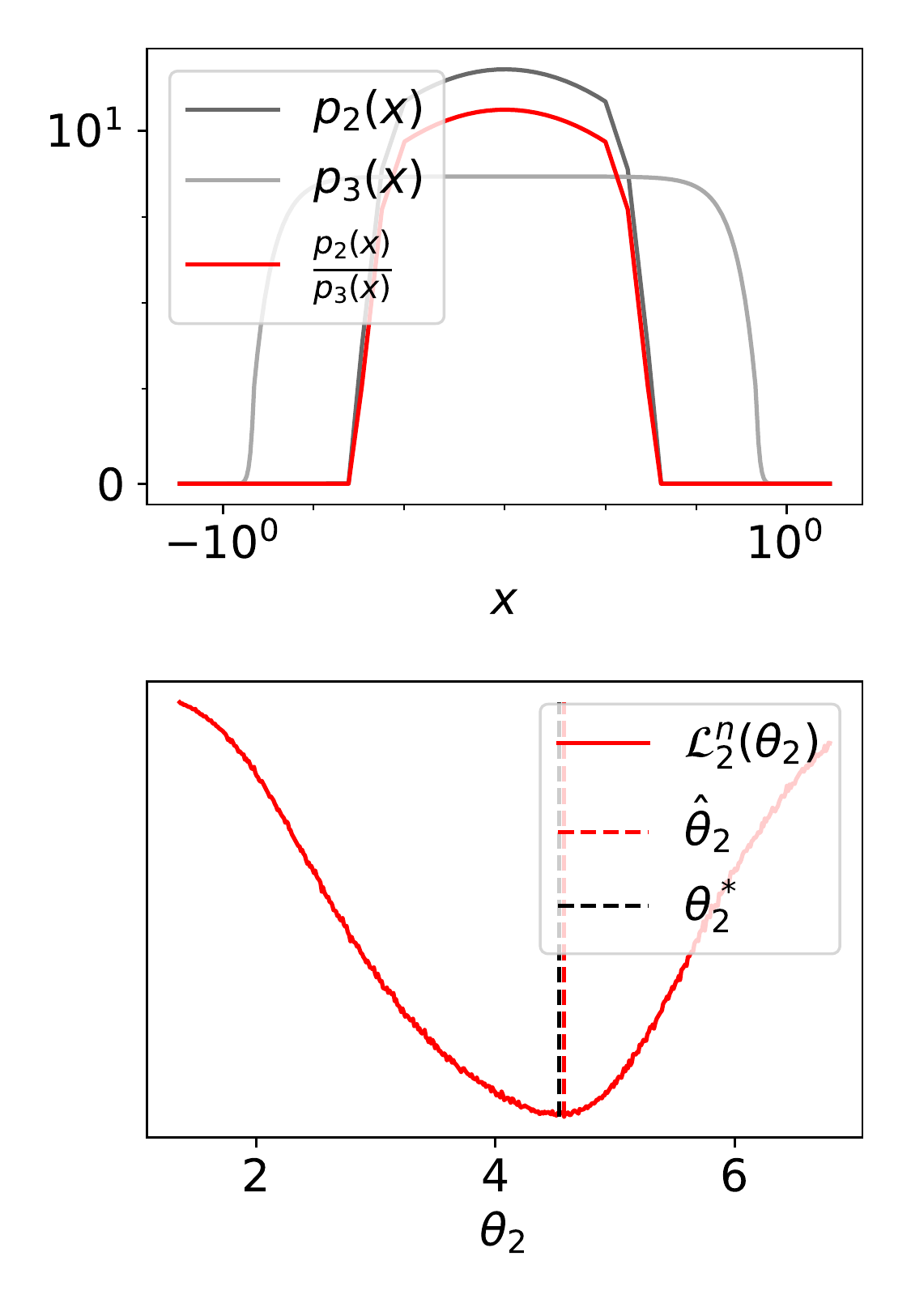} &
        \hspace{-0.5cm}
        \includegraphics[width=.24\linewidth]{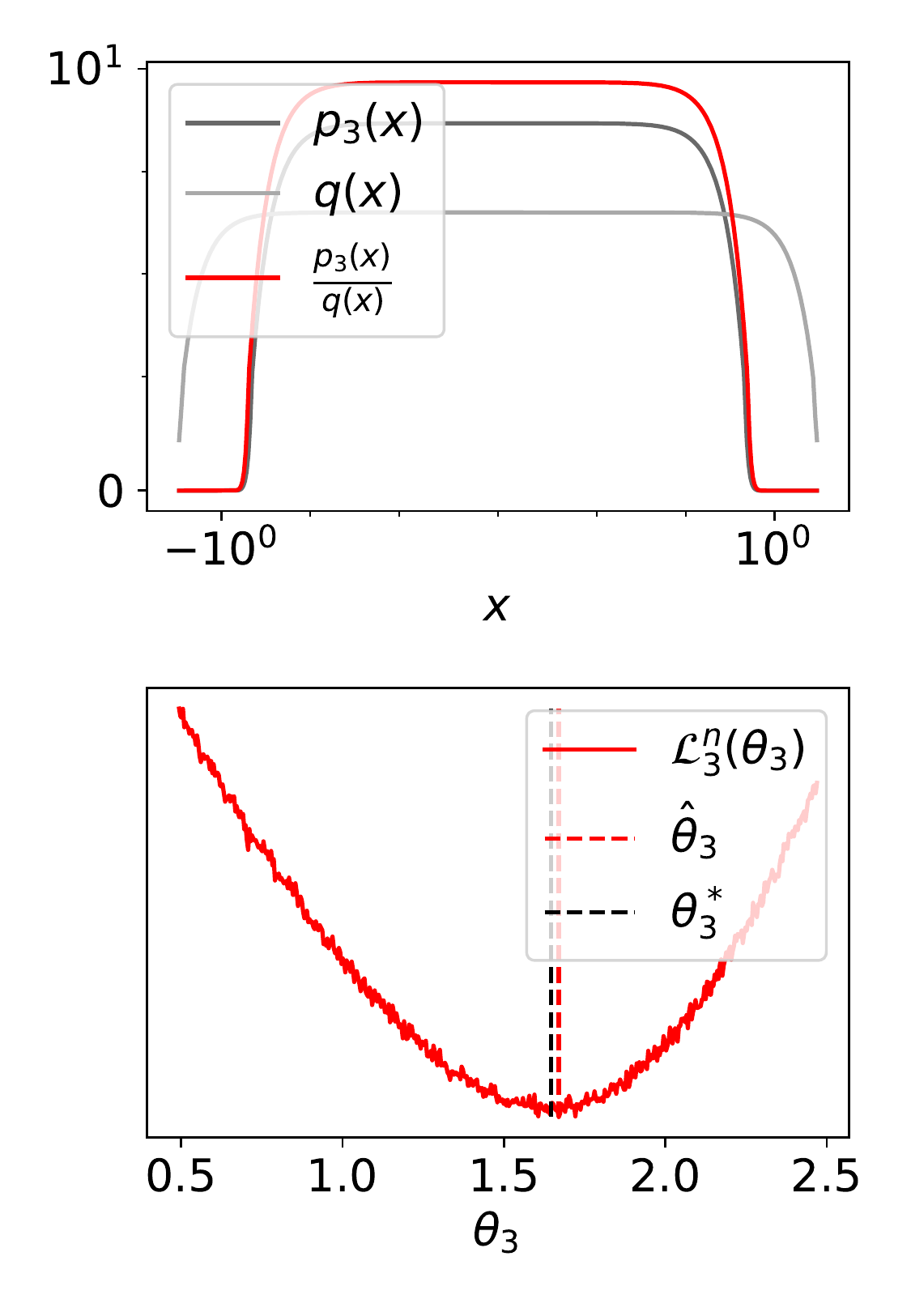}
    \end{tabular}
\label{fig:1d tre lsq subfig}
\end{subfigure}
\caption{Replica of Figure 1 from the main text, except that we use the least-square loss from the GAN literature \cite{mao2017least} for both the single ratio estimator \& for each ratio in TRE.}
\label{fig:1d single ratio vs tre for lsq}
\end{figure}

In Figures \ref{fig:1d single ratio vs tre for nwj} \& \ref{fig:1d single ratio vs tre for lsq}, we can see how single-density ratio estimation performs when using the NWJ and LSQ loss functions for 10,000 samples. the loss curves display the same `saturation' effect seen for the logistic loss, where many settings of the parameter yield an almost identical value of the loss. Moreover, the minimiser of these saturated objectives is far from the `true' minimiser (black dotted lines).

Figures \ref{fig:1d single ratio vs tre for nwj} \& \ref{fig:1d single ratio vs tre for lsq} also show the performance of TRE when each bridge is estimated using the NWJ/LSQ losses. Each TRE loss has a quadratic bowl shape, where the finite-sample minimisers almost perfectly overlap with the true minimisers.

\begin{figure}[h]
\centering
\hspace{0.5cm} \large \textbf{NWJ loss} \hspace{5.5cm} \large \textbf{Least-square loss}
\begin{subfigure}{0.45\textwidth}
    \includegraphics[width=1.0\linewidth]{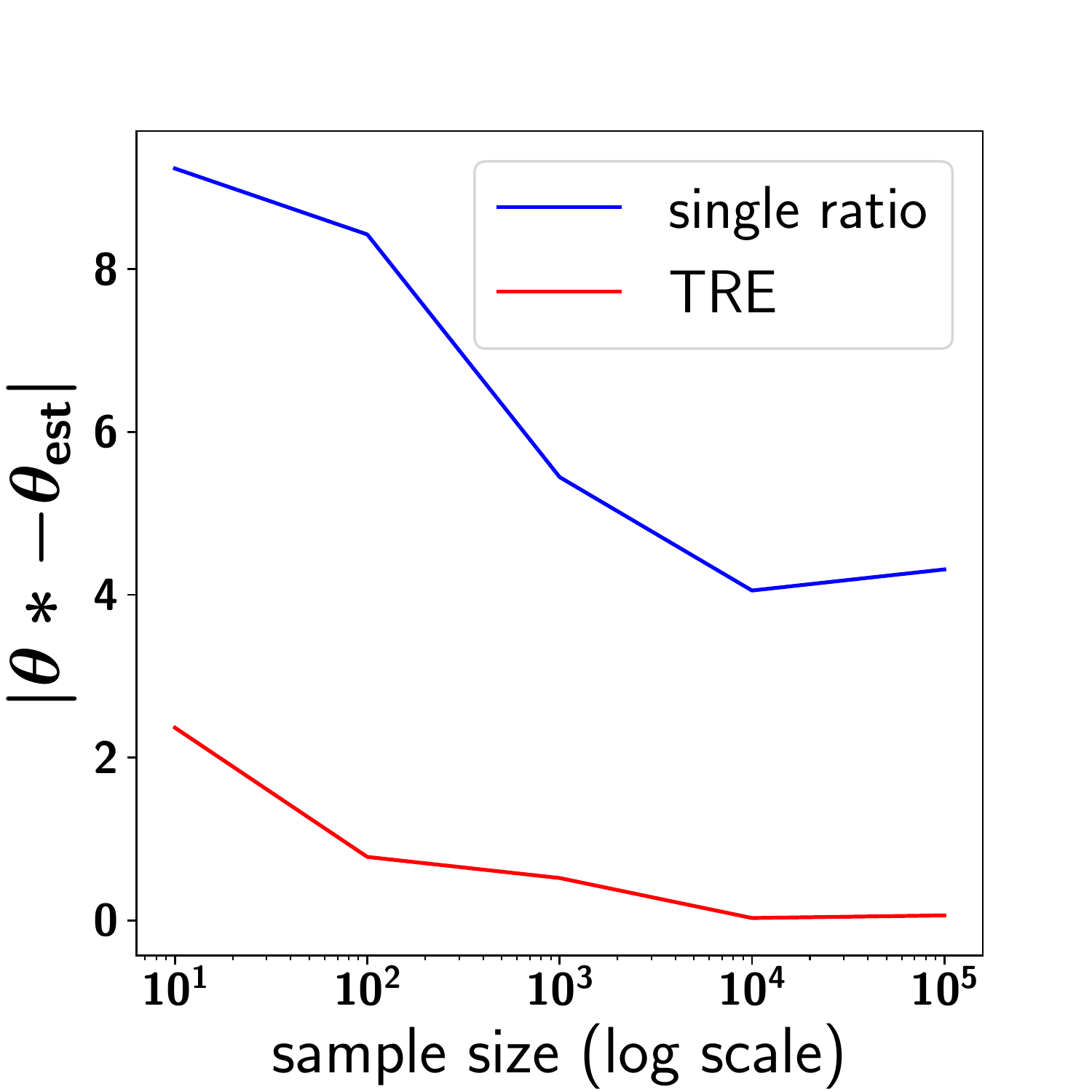} 
\end{subfigure}\hfill
\begin{subfigure}{0.45\textwidth}
    \includegraphics[width=1.0\linewidth]{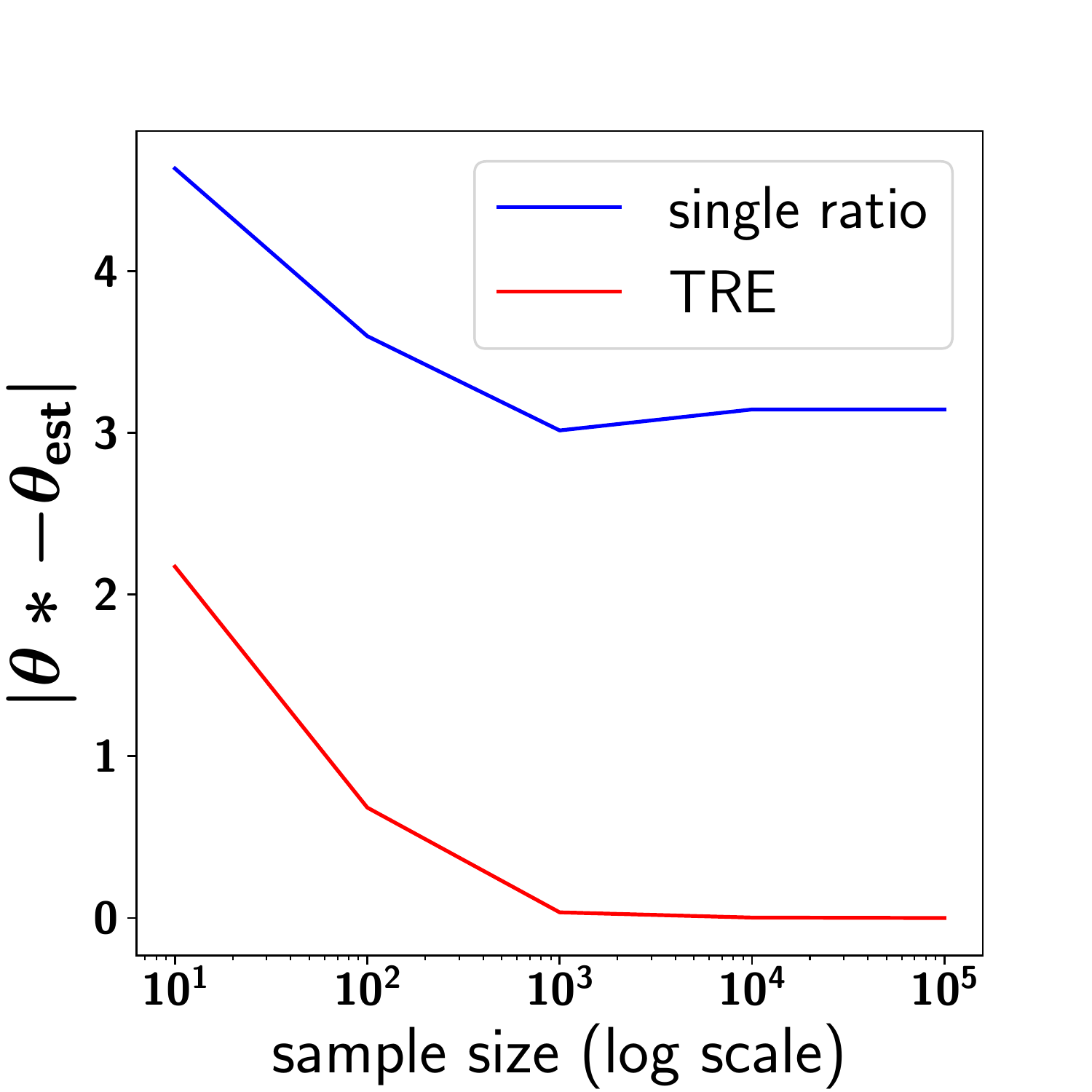} 
\vspace{-0.2cm}
\end{subfigure}
\caption{Sample efficiency curves for the 1d peaked ratio experiment, using different loss functions.}
\label{fig: sample efficiency curves nwj and lsq}
\end{figure}

Finally, we plot sample efficiency curves for both the NWJ and LSQ losses, showing the results in Figure \ref{fig: sample efficiency curves nwj and lsq}. We see that single density-ratio estimation with NWJ or LSQ performs poorly, with at best linear gains for exponential increases in sample size. In contrast, if we perform TRE using NWJ or LSQ losses, then we obtain significantly better performance with orders of magnitude fewer samples. These findings are essentially the same as those presented in the main paper for the logistic loss.

\section{High-dimensional ratio with large MI toy experiment}
In this experiment we estimate the ratio $p_0/p_m$, where both densities are Gaussian, $p_0 = \N(0, \Sigma)$ and $p_m = \N(0, \I)$, where $\Sigma$ is a block-diagonal covariance matrix, where each block is $2\times2$ with 1 on the diagonal and $0.8$ on the off-diagonal. Since we know its analytic form, we can view $p_m$ as a noise distribution, and the ratio-estimation task as an energy-based modelling problem. Alternatively, we may view the problem as a mutual information estimation task, by taking the random variable $\x = (x_1, \ldots, x_d) \sim p_0$, and defining $\u = (x_1, x_3, \ldots, x_{d-1})$ and $\v = (x_2, x_4 \ldots x_d)$. By construction, we therefore have $p(\u)p(\v) = \N(\x; 0, \I) = p_m(\x)$.

We generate $100,000$ samples for each of the train/validation/test splits. We use a total batch size of $1024$, which includes all samples from the waymark trajectories. The bridges in TRE have the form $\log r_k(\x) = \x^T \W_k \x + b_k$, where we enforce that the diagonal entries of $\W_k$ are positive and that the matrix is symmetric. We use the Adam optimiser \cite{kingma2014adam} with an initial learning rate of $0.0001$ for TRE, and $0.0005$ for single ratio estimation. We use the default Tensorflow settings for $\beta_1, \beta_2$ and $\epsilon$. We train the models for $40,000$ iterations, which takes at most 1 hour.

\section{MI estimation \& representation learning on SpatialMultiOmniglot}
We here describe how we created the SpatialMultiOmniglot dataset and give the derivation for the ground truth mutual information values presented in the main paper\footnote{The original work from which we borrow this experiment \cite{ozair2019wasserstein} did not not provide a detailed explanation or code.}. We will share the dataset, along with code for the paper, upon publication. We also state the hyperparameter settings used in our experiments.

\subsection{Dataset construction}
We take the Tensorflow version of the Omniglot dataset (\url{https://www.tensorflow.org/datasets/catalog/omniglot}) and resize it to $28 \times 28$ using the \texttt{tf.image.resize} function. We arrange the data into alphabets $\{A_i\}_{i=1}^{l}$, where each alphabet contains $n_i$ characters. The alphabets are sorted by size, so that $n_1 > n_2 > \ldots > n_l$. Each character in a alphabet has $20$ different \emph{versions} (e.g.\ there are 20 different images depicting the letter `w'). Hence, we can express each alphabet as a set $A_i = \{ \{ a_{j, k}^i \}_{k=1}^{20} \}_{j=1}^{n_i}$, where $a_{j, k}^i$ refers to the $k$\textsuperscript{th} version of the $j$\textsuperscript{th} character of the $i$\textsuperscript{th} alphabet.

In order to construct the $d$-dimensional version of the SpatialMultiOmniGlot dataset, we restrict ourselves to the $d$ largest alphabets $\{A_i\}_{i=1}^{d}$. We then sample a vector of categorical random variables
\begin{align}
    \j = (j_1, \ldots, j_d) \sim \text{Cat}(n_1) \times \ldots \text{Cat}(n_d) 
\end{align}
where the $i$\textsuperscript{th} categorical distribution is uniform over the set $\{1, \ldots, n_i\}$ and is independent from the other categorical distributions. The vector $\j$ should be thought of as an index vector that specifies a particular character from each of the $d$ alphabets.

We then sample two i.i.d random variables $\k$ and $\k'$, via
\begin{align}
    \k &= (k_1, \ldots, k_d) \sim \prod_{i=1}^d \text{Cat}(20) & \k' = (k'_1, \ldots, k'_d) \sim \prod_{i=1}^d \text{Cat}(20)
\end{align}
where, again, each Categorical distribution is independent from the rest. These vectors should be thought of as index vectors that specify a particular version of a character.

Now, we define a datapoint as a tuple $\x = (\u, \v)$, where
\begin{align}
    \u &= (a^1_{j_1, k_1}, \ldots, a^d_{j_d, k_d}) &
    \v = (a^1_{j_1+1, k'_1}, \ldots a^d_{j_d+1, k'_d}).
\end{align}
In words, we construct $\u$ and $\v$ such that $u_i$ and $v_i$ are consecutive characters within their alphabet (whilst the precise \emph{versions} of the characters are randomised). Finally, we arrange $\u$ and $\v$ into a grid using raster ordering. This is possible since we assume $d$ to be a square number.

Importantly, we emphasise that $\u, \v \in \prod_{i=1}^{d} A_i$ are discrete random variables defined over a set of template images. They are not defined over a space of pixel values, as is usually the case in image-modelling.

\subsection{Derivation of ground truth MI}
By construction, we have that $\u$ and $\v$ are conditionally independent given $\j$. This means
\begin{align}
\label{eq: omniglot conditional independence}
    p(\u \lvert \ \v, \j) = p(\u \lvert \ \j).
\end{align}
Furthermore, will assume that, for all $\u$ there exists a unique $\j_{\u}$ such that
\begin{align}
 \label{eq: omniglot unique j given u}
     p(\j_{\u} \lvert \ \u) =  1.
\end{align}
Similarly, for any $\v$, there exists a unique $\j_{\v}$ satisfying the same condition. In words, this simply means that, given a grid of Omniglot images, we assume there is no ambiguity about which characters are present. Using Bayes' rule, and the fact that for a given $\j$, $\u$ is uniquely determined by $\k$, one can then deduce that
\begin{align}
    p(\u \lvert \ \j) = \left\{\begin{array}{lr}
        0, & \text{for } \j \neq \j_{\u} \\
        20^{-d}, & \text{for } \j = \j_{\u} \\
        \end{array}\right\}. \label{eq: omniglot u given j}
\end{align}
and similarly for $\v$.

We now proceed to derive an analytical formula for the ground truth mutual information between $\u$ and $\v$. We show that the mutual information is equal to the sum of the log alphabet sizes $\mathcal{I}(\u, \v) = \sum_{i=1}^d \log n_i$.

This holds because
\begin{flalign}
    \mathcal{I}(\u, \v) &= \E_{p(\u, \v)} \log \frac{p(\u, \v)}{p(\u)p(\v)} & \\[1.5pt]
    &= \E_{p(\u, \v)} \log \frac{p(\u \lvert \ \v)}{p(\u)} & \\[1.5pt]
    &= \E_{p(\u, \v)} \log \frac{ \sum_{\j} p(\u, \j \lvert \ \v)}{ \sum_{\j} p(\u, \j)} & \\[1.5pt]
    &= \E_{p(\u, \v)} \log \frac{ \sum_{\j} p(\u \lvert \ \j) p(\j \lvert \ \v)}{ \sum_{\j} p(\u \lvert \ \j)p(\j)} \hspace{1cm} \text{by \eqref{eq: omniglot conditional independence}} & \\[1.5pt]
    &= \E_{p(\u, \v)} \log \frac{ p(\u \lvert \ \j_{\v})}{ \sum_{\j} p(\u \lvert \ \j)p(\j)} \hspace{1.4cm} \text{by \eqref{eq: omniglot unique j given u}} & \\[1.5pt]
    &= \E_{p(\u, \v)} \log \frac{ p(\u \lvert \ \j_{\v})}{ p(\u \lvert \ \j_{\u}) p(\j_{\u})} \hspace{1.6cm} \text{by \eqref{eq: omniglot u given j}} & \\[1.5pt]
    &= \E_{p(\u, \v)} \log \frac{1}{p(\j_{\u})} \hspace{2.7cm} \text{since } \j_{\v} = \j_{\u} \text{ if } p(\u, \v) > 0 & \\[1.5pt]
    &= \E_{p(\u, \v)} \log \big( \prod_{i=1}^{d} n_i \big) \hspace{2.3cm} \text{since } p(\j) \text{ is uniform}  & \\[1.5pt]
    &= \sum_{i=1}^d \log n_i
\end{flalign}

\subsection{Experimental settings}
We generate 3 versions of the SpatialMultiOmniglot dataset for $d = 1, 4, 9$. For each version, we sample $50,000$ training points and $10,000$ validation and test points. As stated in the main text, we use a separable architecture given by 
\begin{align}
    \log r_k(\u, \v) = g(\u)^T \W_k f_k(\v),
\end{align}
where $f_k$ is a convolutional ResNet whose architecture is given in Figure \ref{table: multiomniglot architecture}. The function $g$ is also a convolutional ResNet with almost the same architecture, except that none of its parameters are bridge-specific, and hence the `ConditionalScaleShift' layers simply become `ScaleShift' layers, with no dependence on $k$.

To construct a mini-batch, we first sample a batch from the joint distribution $p(\u, \v)$. We then obtain samples from $p(\u)p(\v)$ by sampling a \emph{second} batch from the joint distribution (which could overlap with the first batch), and shuffling the $\v$ vectors across this second batch. Finally, we construct waymark trajectories as described in the main text. For all experiments, the `total' batch size is $\sim 512$, which includes all samples from the waymark trajectories. Thus, as the number of waymarks increases, the number of trajectories in a batch decreases.

We use the Adam optimiser \cite{kingma2014adam} with an initial learning rate of $10^{-4}$ with default Tensorflow settings for $\beta_1, \beta_2$ and $\epsilon$. We gradually decrease the learning rate over the course of training with cosine annealing \cite{Loshchi2017cosine}. All models are trained using a single NVIDIA Tesla P100 GPU card for $200,000$ iterations, which takes at most a day.

We grid-searched over the type of pooling (max vs.\ average) and the size of the final dense layer ($150n, 300n$ and $450n$, where $d = n^2$). Interestingly, average pooling was less prone to overfitting and often yielded better final performance, however it was often `slow to get started', with the TRE losses hardly making any progress during the first quarter of training.

For the representation learning evaluations, we first obtained the hidden representations $g(\u)$ for the entire dataset. We then trained a collection of \emph{independent} supervised linear classifiers on top of these representations, in order to predict the alphabetic position of each character in $\u$. We used the L-BFGS optimiser to fit these classifiers via the \texttt{tfp.optimizer.lbfgs\_minimize} function, setting the maximum iteration number to $10,000$.

\section{Energy-based modelling on MNIST}
We here discuss the parameterisation of the noise distributions used in the experiments, the exact method for sampling from the learned EBMs, and the experimental settings used for TRE.

For all noise distributions and TRE models, we use the Adam optimiser \cite{kingma2014adam} with an initial learning rate of $10^{-4}$ with default Tensorflow settings for $\beta_1, \beta_2$ and $\epsilon$. We gradually decrease the learning rate over the course of training with cosine annealing \cite{Loshchi2017cosine}. All models are trained using a single NVIDIA Tesla P100 GPU card.

\subsection{Noise distributions}
As stated in the main text, we consider three noise distributions: a multivariate Gaussian, a Gaussian copula and a rational-quadratic neural spline flow (RQ-NSF), all of which are pre-trained via maximum likelihood estimation.

The full-covariance multivariate Gaussian is by far the simplest, and can be fitted in around a minute via \texttt{np.cov}. The Gaussian copula is slightly more complicated. Its density can be written as 
$p(\x) = \N( [s_1(x_1), \ldots, s_d(x_d)]; \mu, \Sigma) \prod_{i=1}^d |s'_i(x_i)|$. The $s_i$ are given by the composition of the inverse CDF of a standard normal and the CDF of the univariate $x_i$. It is possible to exploit this to learn the $s_i$---as well as $\mu$ and $\Sigma$---however, we found it slightly simpler to directly parametrise the $s_i$ via flexible rational-quadratic spline functions \cite{durkan2019neural} of which there are official implementations in Tensorflow and Pytorch and to jointly learn all parameters via maximum likelihood. We follow the basic hyperparameter recommendations in \cite{durkan2019neural}. The hyperparameters that required tuning were the number of bins (we use 128) and the interval widths (which we set to 3 times the standard deviation of the data). For optimisation, we used a batch size of 512 and trained for $40,000$ iterations.

Finally, we turn to the RQ-NSF model \cite{durkan2019neural}. We largely adopt the architectural choices of \cite{durkan2019neural}, and so for a more detailed explanation, we refer the reader to their work. We use a multi-scale convolutional architecture comprised of 2 levels, where each level contains 8 `steps'. A step consists of an actnorm layer, an invertible $1 \times 1$ convolution, and a rational-quadratic coupling transform. The coupling transforms are parameterised by a block of convolution operations following \cite{Kumar2020VideoFlow}, which use 64 feature maps. The spline functions use 8 bins and the interval width is set to [-3, 3]. We do not `factor out' half of the variables at the end of each level, but do perform `squeeze' operation and an additional $1 \times 1$ convolution. For optimisation, we set the batch size to 256, the dropout rate to 0.1, and train for $200,000$ iterations, which takes under a day.

\subsection{Annealed MCMC Sampling}
We here describe how we leverage the specific  products-of-experts structure of the TRE model to perform annealed MCMC sampling. Firstly, we initialise a set of MCMC chains with i.i.d samples from the noise distribution $p_m$. We could then run an MCMC sampler with the full TRE model as the target distribution. However, we instead use an annealing procedure, whereby we iteratively sample from a sequence of distributions that interpolate between $p_m$ and $p_0$. Such distributions can be obtained by multiplying $p_m$ with an increasing number of bridges
\begin{align}
    & p_k(\x) = p_m(\x) \prod_{i=k}^{m-1} r_k(\x), & k = m-1, \ldots 0. \label{eq: intermediate annealing dists}
\end{align}
To obtain an even smoother interpolation, we further define exponentially-averaged intermediate distributions $p_{k, t}(\x) = p_k(\x)^{\beta_t} p_{k+1}(\x)^{1 -\beta_t}$, where $\{\beta_t\}$ is a decreasing sequence of numbers ranging from 1 to 0.

In addition to obtaining samples, we can simultaneously use this annealing procedure for estimating the log-likelihood of the model via annealed importance sampling (AIS) \cite{neal2001annealed}. We may also run the annealing procedure `in reverse', initialising a chain at a datapoint and iteratively removing bridges until the target distribution of the MCMC sampler is the noise distribution. Using this reverse sampling procedure, we can obtain a second, more conservative, estimate of the log-likelihood via the reverse annealed importance sampling estimator (RAISE) \cite{burda2015accurate}.

Whilst in principle any MCMC sampler could be used, the efficiency of different samplers can vary greatly. We choose to use the gradient-based No-U-turn sampler (NUTS) \cite{hoffman2014no}, which is a highly efficient method for many applications. We use the official Tensorflow implementation along with most of the default hyperparameter settings. We set the target acceptance rate to 0.6, and use a max tree depth of 6 during the annealed sampling. We also continue to run the sampler after the annealing phase is finished, using a max tree depth of 10. We use a total of $1000$ intermediate distributions with $100$ parallel chains.

Finally, recall from the main text that each noise distribution in our experiments can be expressed as invertible transformation $F$ of a standard normal distribution. We use this $F$ to further enhance the efficiency of the NUTS sampler, by performing the sampling in the $\z$-space, and then mapping the final results back to $\x$-space. Working in $\z$-space, by the rules of transformations of random variables, the intermediate distributions of \eqref{eq: intermediate annealing dists} become
%$p_k(\z) = \mathcal{N}(\z; 0, \I) \prod_{i=k}^{m-1} %r_k(F(\z))$.
\begin{align}
    p_k(\z) = \mathcal{N}(\z; 0, \I) \prod_{i=k}^{m-1} r_k(F(\z)). 
\end{align}

AIS and RAISE can still be applied, just as before, to obtain an estimate of the log-likelihood in $\z$-space. The change of variables formula for probability density functions can then be applied to obtain estimated log-likelihoods for the original TRE model in $\x$-space. We note that when the noise distribution is a normalising flow, prior work has demonstrated that $\z$-space MCMC sampling can be significantly more effective than working in the original data space \cite{hoffman2019neutra}.

\subsection{Experimental settings}
We use the standard version of the MNIST dataset \cite{lecun1998mnist}, with $50,000$ training points, and $10,000$ validation and test points. We follow the same preprocessing steps as \cite{papamakarios2017masked}, `dequantizing' the dataset with uniform noise, re-scaling to the unit interval, and then mapping to the real line via a logit transformation.

The architecture for the TRE bridges is given in Figure \ref{table: mnist architecture}. The waymark mechanism and associated grid-search is given in Table \ref{table: wmark spacing}. A consistent observation across all our MNIST experiments was that the first ratio-estimator between the data distribution $p_0$ and a slightly perturbed data distribution $p_1$ was extremely prone to overfitting. We found that the only way to mitigate this problem was to simply drop the ratio by setting the $\alpha_0$ in \eqref{eq: appendix linear combination waymarks} to a very small value (0.01) rather than exactly 0. Equivalently, this can be viewed as applying standard TRE to a very slightly perturbed data distribution. We note that this perturbation is small enough that is barely visible to the human eye when comparing samples. We conjecture that this problem may stem from the fact that the original MNIST dataset is actually discrete not continuous and the `dequantizing' perturbation used to make the data continuous is perhaps not sufficient.

To form mini-batches, we sample $25$ datapoints each from $p_0$ and $p_m$, and then generate waymark trajectories as described in the main text. Thus, the total batch size is $25 \times (m+1)$. We use the optimisation settings described at the beginning of this section, training for $200,000$ iterations, which takes about a day.

\subsection{Additional results}
\begin{figure}[b]
    \centering
    \includegraphics[width=0.95 \textwidth]{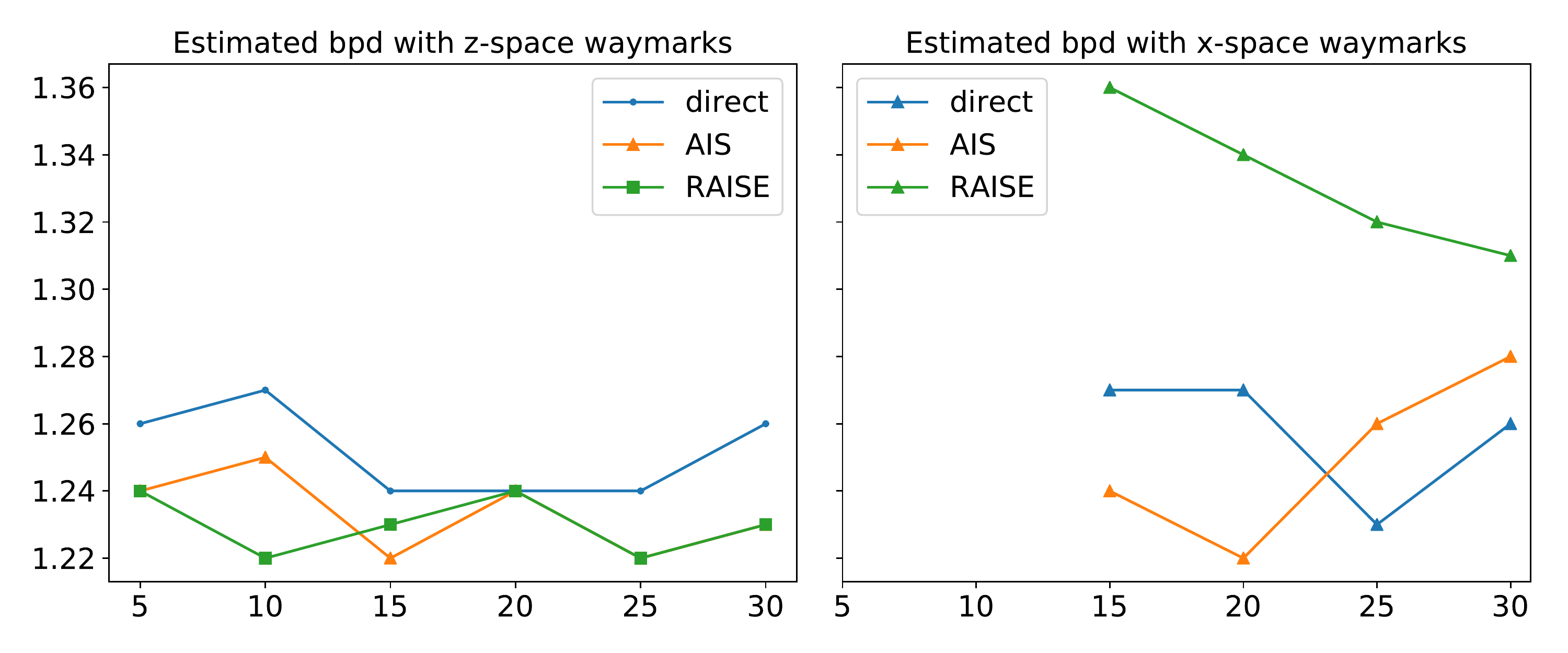}
    \caption{Waymark sensitivity analysis for TRE with copula noise distribution. Both plots show the estimated bits per dimension (bpd) as a function of waymark number. On the left we apply the linear combination waymark mechanism in $\z$-space, whilst on the right we apply it in $\x$-space. As described in Section \ref{sec: wmark spacing}, we terminate runs where any of the TRE losses saturate close to 0, which is exactly what happened when using the $\x$-space mechanism for $5$ and $10$ waymarks.}
    \label{fig: mnist wmark sensitivity}
\end{figure}

In Figure \ref{fig: mnist wmark sensitivity}, we present a sensitivity analysis showing how the quality of the learned EBM varies as we alter the number of waymarks, as well as the space in which the waymarks are generated. We found that working in $\x$-space yielded lower performance compared to working in $\z$-space, as measured by the most conservative estimator, RAISE. In particular, we found that the $\x$-space mechanism required more waymarks (around 15) to avoid any of the logistic losses saturating close to 0, and it was significantly harder to tune the spacing of the waymarks as indicated by Table \ref{table: wmark spacing}.

\begin{figure}
 \centering
    \begin{subfigure}{0.8\textwidth}
          \centering
          \includegraphics[width=0.8
          \textwidth]{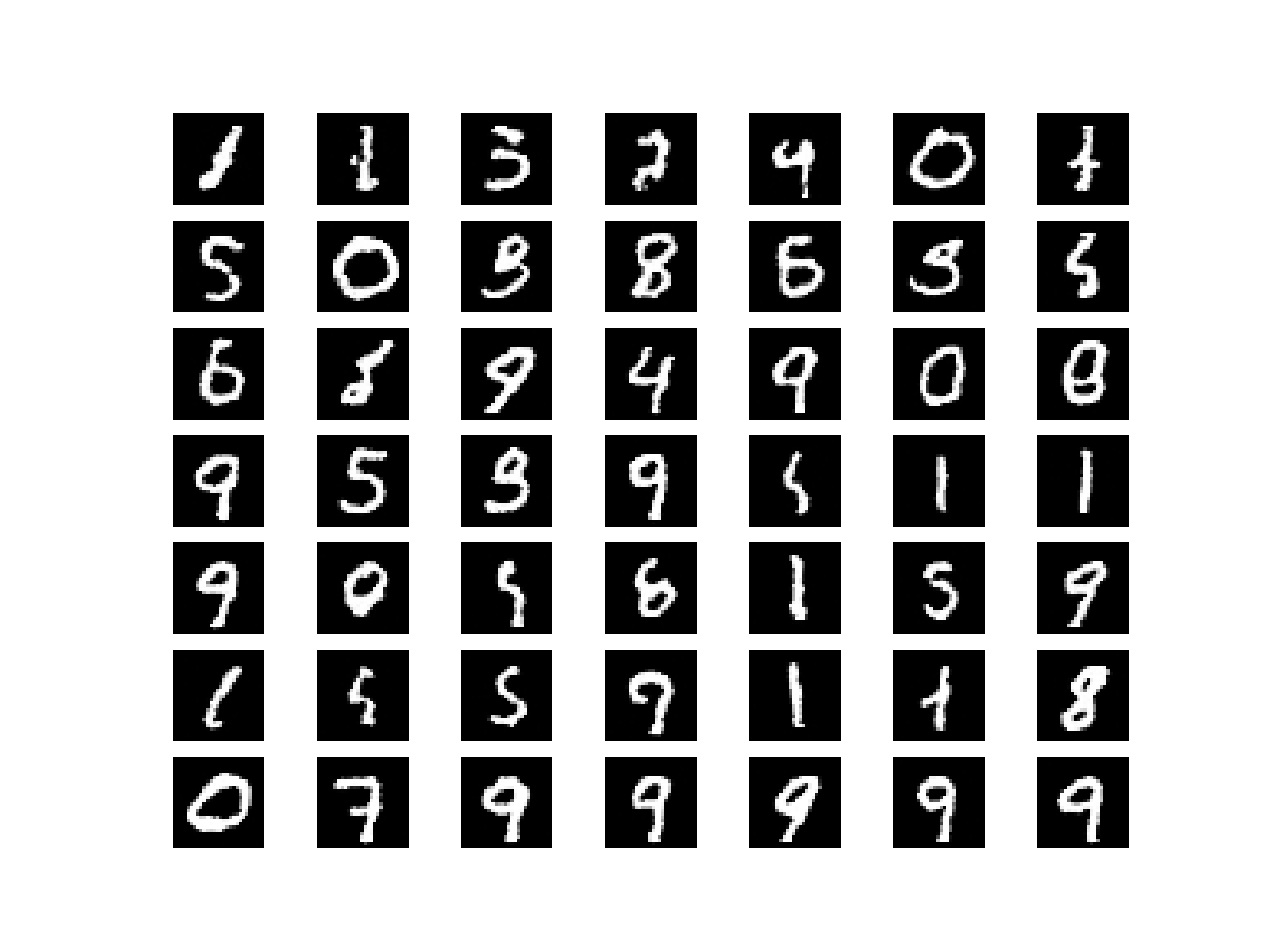}
        %   \vspace{-0.5cm}
          \caption{Samples from TRE model with Gaussian noise distribution, ordered by log-density.}
    \end{subfigure}
        \begin{subfigure}{0.8\textwidth}
          \centering
            \includegraphics[width=0.8 \textwidth]{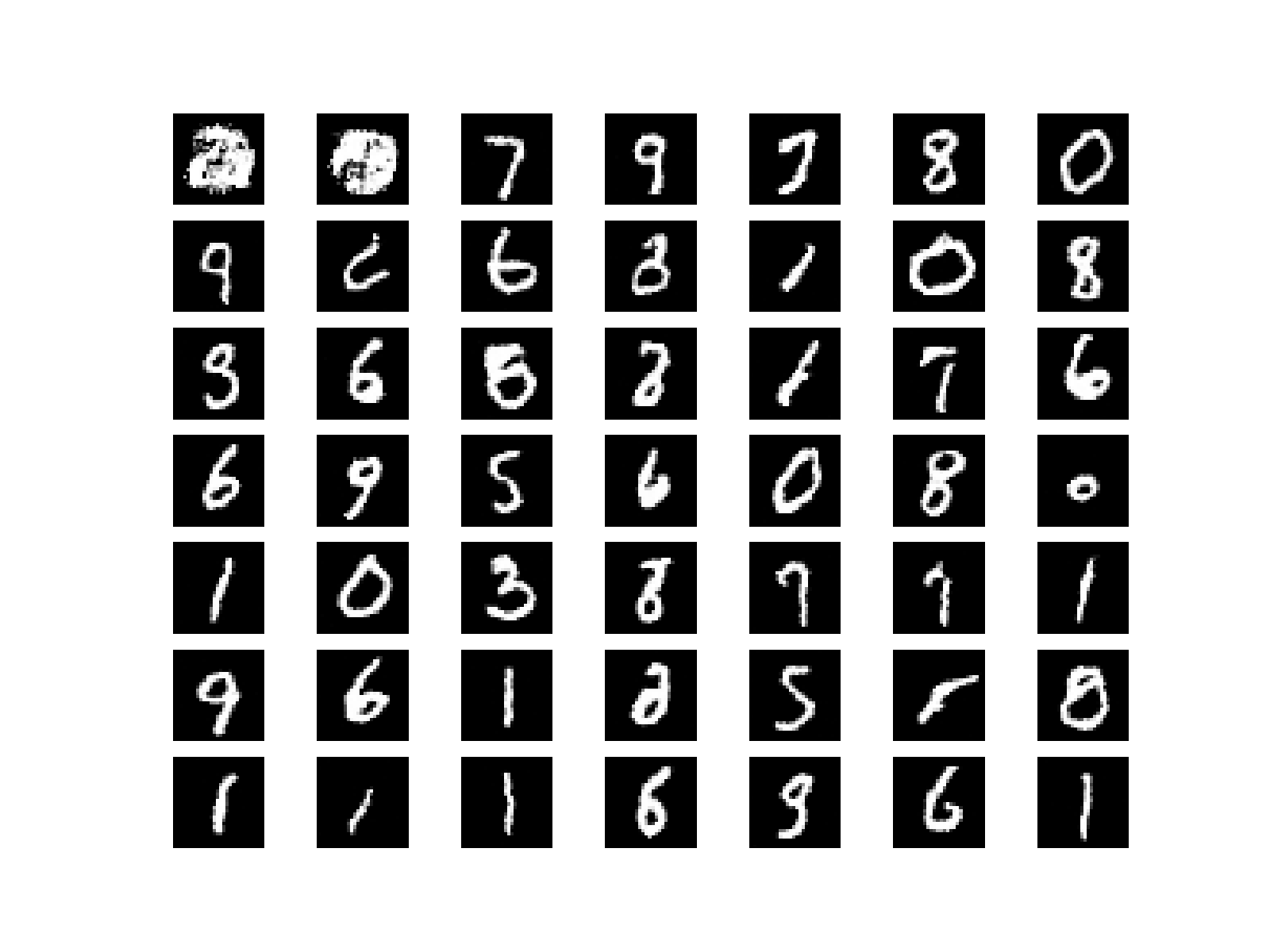}
            % \vspace{-0.5cm}
            \caption{Samples from TRE model with copula noise distribution, ordered by log-density.}
    \end{subfigure}
        \begin{subfigure}{0.8\textwidth}
          \centering
            \includegraphics[width=0.8 \textwidth]{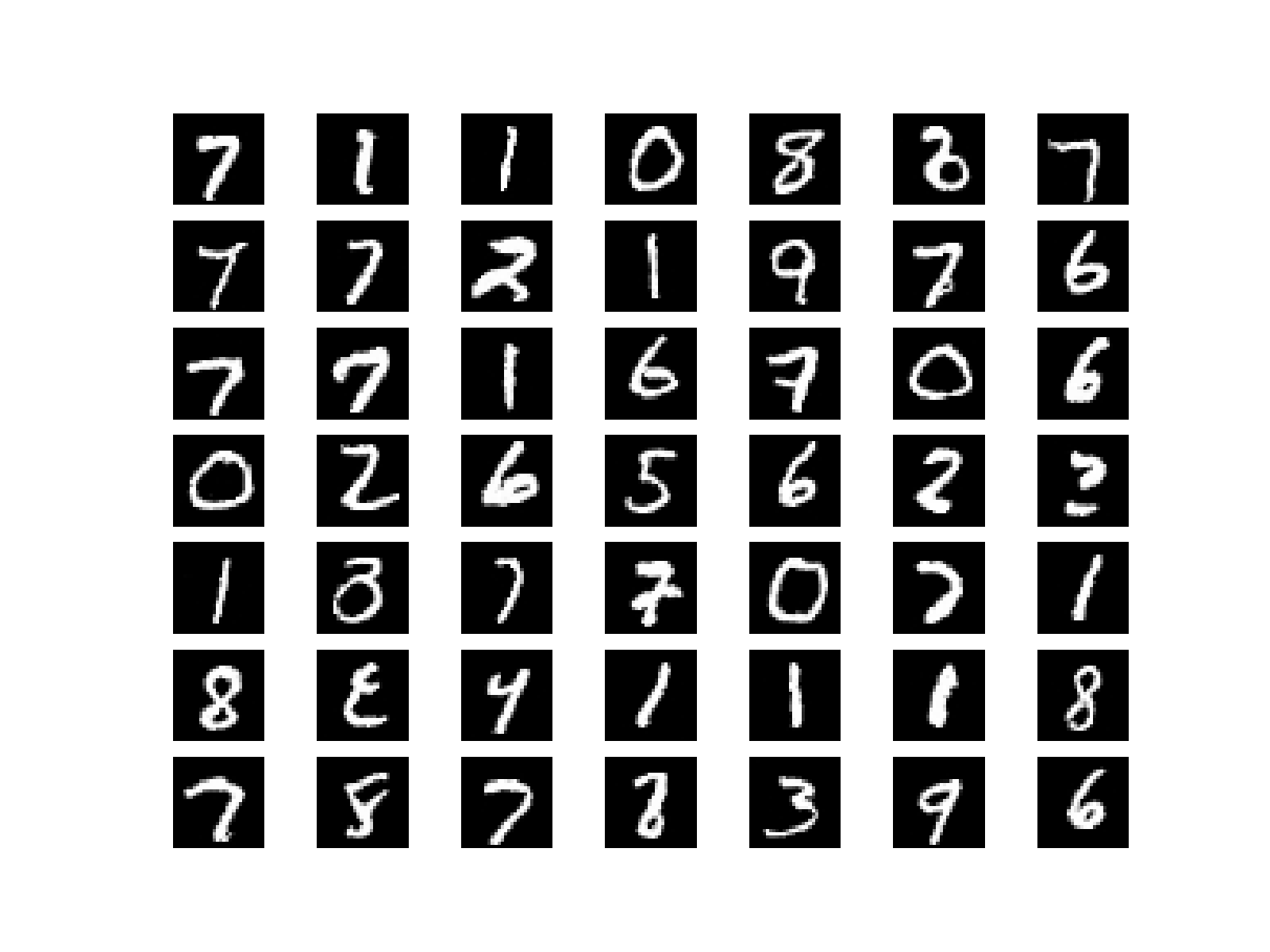}
        % \vspace{-0.5cm}
        \caption{Samples from TRE model with RQ-NSF noise distribution, ordered by log-density.}
    \end{subfigure}
    \caption{Extended MNIST samples}
    \label{fig: extended mnist samples}
\end{figure}

Finally, for the models whose results were given in the main paper, we display extended image samples in Figure \ref{fig: extended mnist samples}. Note that these samples are ordered by log-density (lowest density in top left corner, highest in bottom right).

\end{appendix}

\end{document}